%% file: egpaper.tex
\documentclass[10pt,twocolumn,letterpaper]{article}

\usepackage{wacv}              
\usepackage[accsupp]{axessibility}  

\usepackage{graphicx}
\usepackage{amsmath}
\usepackage{amssymb}
\usepackage{booktabs}
\usepackage{times}
\usepackage{epsfig}
\usepackage{xspace}
\usepackage{comment}
\usepackage{color}
\usepackage[accsupp]{axessibility}  
\usepackage{caption}
\usepackage{subcaption}
\usepackage{url}
\usepackage{tabularx} 
\usepackage{multirow}
\usepackage{tikz}
\usetikzlibrary{tikzmark}
\usepackage[
  separate-uncertainty = true,
  multi-part-units = repeat
]{siunitx}
\usepackage{enumitem}

\usepackage{acronym}
\newacro{DGM}{deep generative model}
\newacro{GAN}{generative adversarial network}
\newacro{DPM}{diffusion probabilistic model}
\newacro{VAE}{variational autoencoder}
\newacro{PBR}{physically based rendering}
\newacro{PG}{Pretrained Guidance}
\newacro{RG}{Real Guidance}

\usepackage[pagebackref,breaklinks,colorlinks]{hyperref}

\usepackage[capitalize]{cleveref}
\crefname{section}{Sec.}{Secs.}
\Crefname{section}{Section}{Sections}
\Crefname{table}{Table}{Tables}
\crefname{table}{Tab.}{Tabs.}

\def\wacvPaperID{243} 
\def\confName{WACV}
\def\confYear{2024}

\include{figures}
\include{tables}

\begin{document}
\include{main2}
\clearpage
\include{bibcommand}
\clearpage
\include{appendix}

\clearpage

\end{document}

%% file: figures.tex
\newcommand{\figteaser}{
\begin{figure}[!htbp]
\centering
\includegraphics[width=0.9\linewidth]{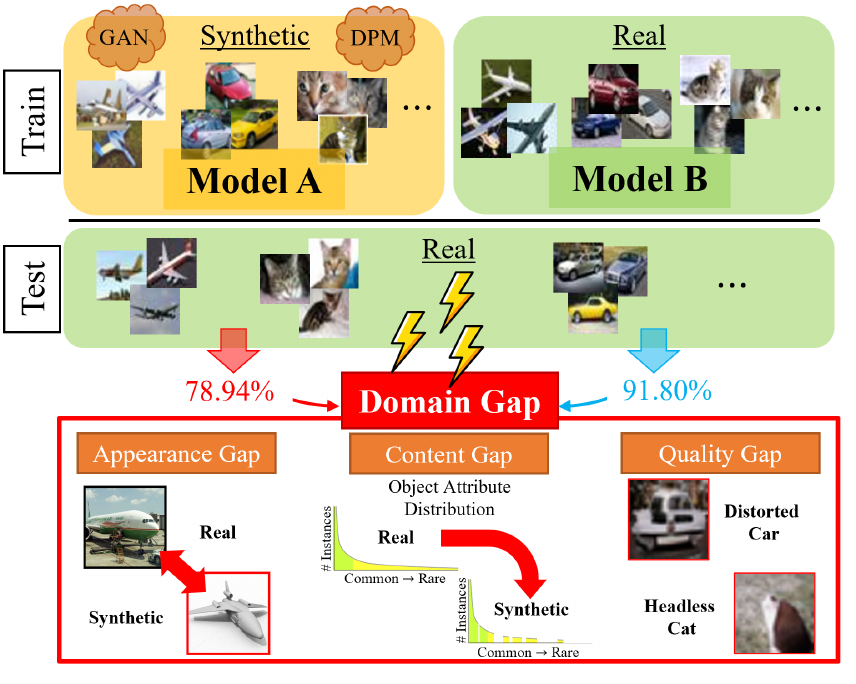}\\
\vspace{-0.2cm}
\caption{An illustration of the commonly observed performance degradation when training a model on synthetic data and testing it on real data. Such effects can be 
attributed to the \emph{Domain Gap}, which we split into three aspects (appearance, content, quality).}
\label{fig:teaser}
\vspace{-0.5cm}
\end{figure}

}

\newcommand{\figtraincurves}{
\begin{figure}[!htbp]
\vspace{-0.3cm}
\centering
\includegraphics[width=0.9\linewidth]{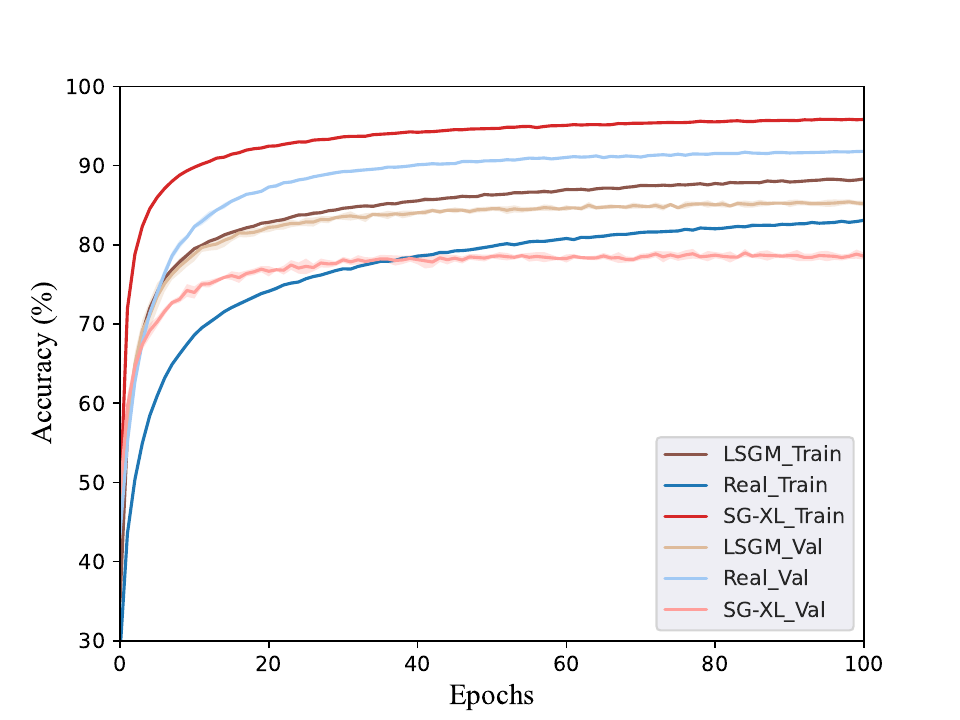}\\
\vspace{-0.3cm}
\caption{The training and validation curves of CIFAR-10 images from different sources over five runs. Standard deviations are plotted as shaded areas. (Zoom in.)}
\label{fig:training-curves}
\vspace{-0.4cm}
\end{figure}
}

\newcommand{\figfakelossdistr}{
\begin{figure*}[!tbp]
\begin{subfigure}[b]{0.5\textwidth}
\centering
\includegraphics[width=0.8\textwidth]{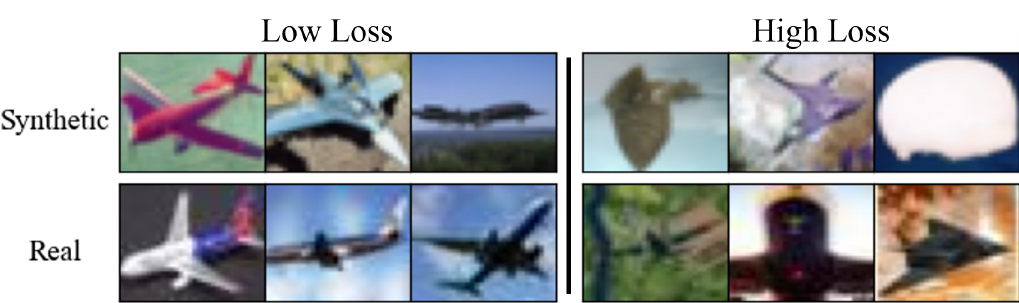}
\includegraphics[width=0.8\textwidth]{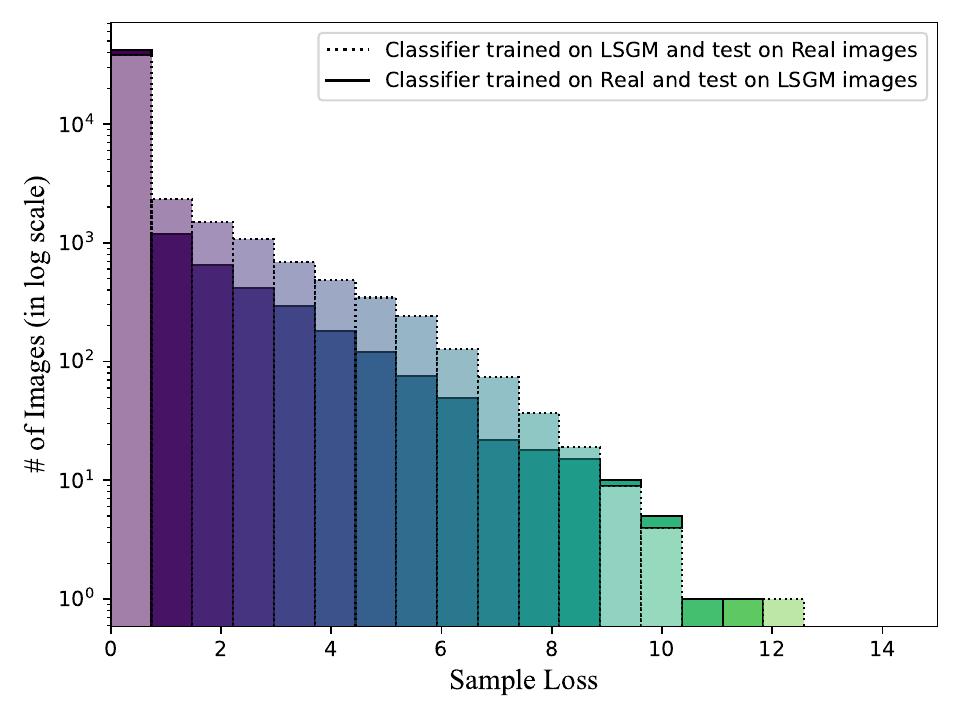}
\vspace{-0.3cm}
\caption{LSGM}
\end{subfigure}
\begin{subfigure}[b]{0.5\textwidth}
\centering
\includegraphics[width=0.8\textwidth]{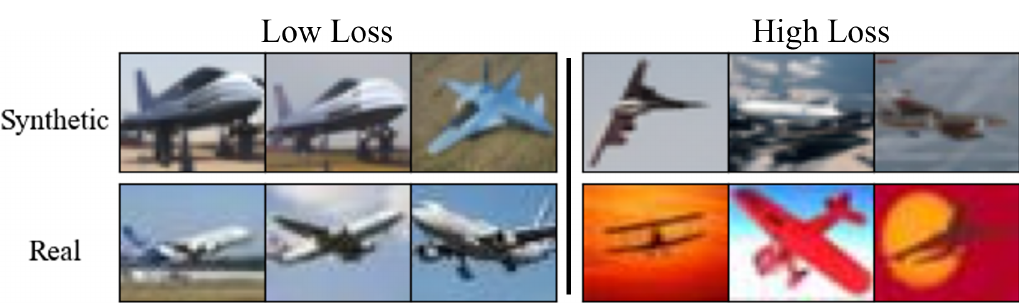}
\includegraphics[width=0.8\textwidth]{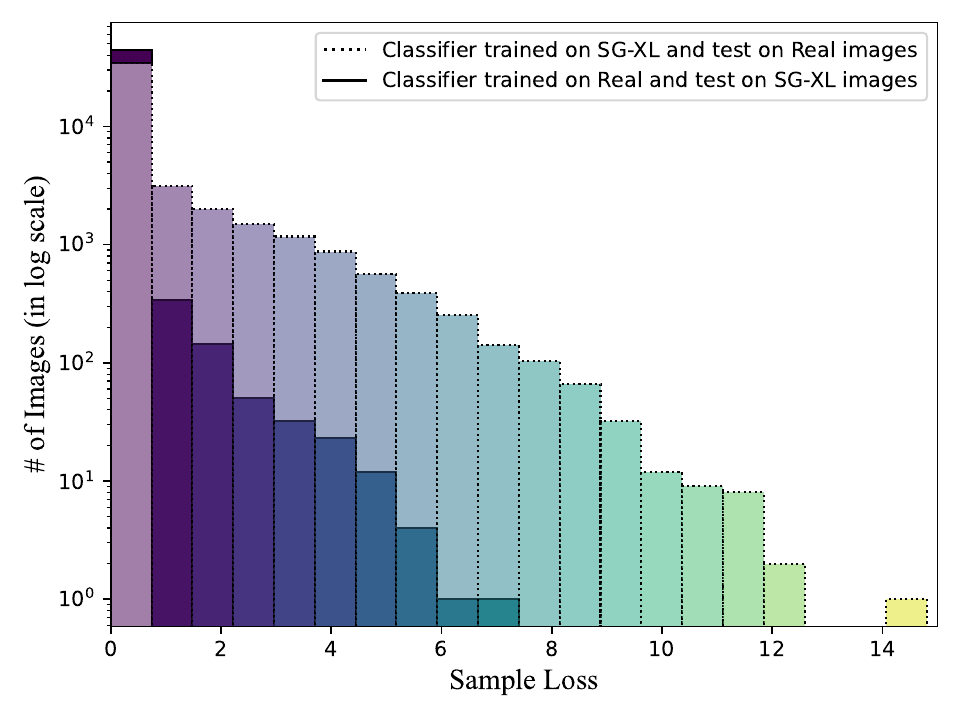}
\vspace{-0.3cm}
\caption{SG-XL}
\end{subfigure}
\vspace{-0.7cm}
\caption{The loss distribution of the samples from different sources.  We overlapped the evaluation of   \emph{Real} $\rightarrow$ \emph{Synthetic} (Solid) and   \emph{Synthetic} $\rightarrow$ \emph{Real} (Dotted) in each subgraph and visualized the associated low-loss and high-loss samples of class Airplane. Note that all models used for evaluation were initialized with ImageNet weights.}
\label{fig:fake-loss-distr}
\end{figure*}
}
\newcommand{\figpgrg}{
\begin{figure*}[!htbp]
\begin{subfigure}[b]{1.0\textwidth}
\centering
\includegraphics[width=0.9\linewidth]{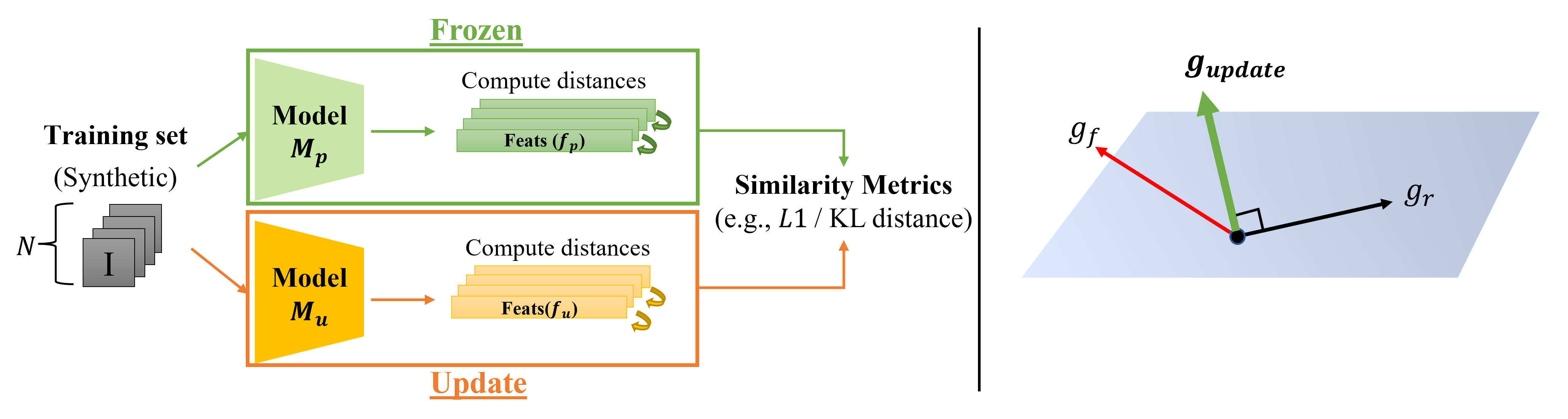}\\
\vspace{-0.3cm}
\caption{Pretrained Guidance \qquad\qquad\qquad\qquad\qquad\qquad\qquad\qquad\qquad\qquad (b) Real Guidance \!\!\!\!\!\!\!\!\!\!\!\!\!\!\!\!} 
\end{subfigure}
\vspace{-0.7cm}
\caption{The proposed \textbf{Pretrained Guidance} and \textbf{Real Guidance}.}
\label{fig:pgrg}
\end{figure*}
}

\newcommand{\figtraincurvesRand}{
\begin{figure}[!tbp]
\centering
\includegraphics[width=0.9\linewidth]{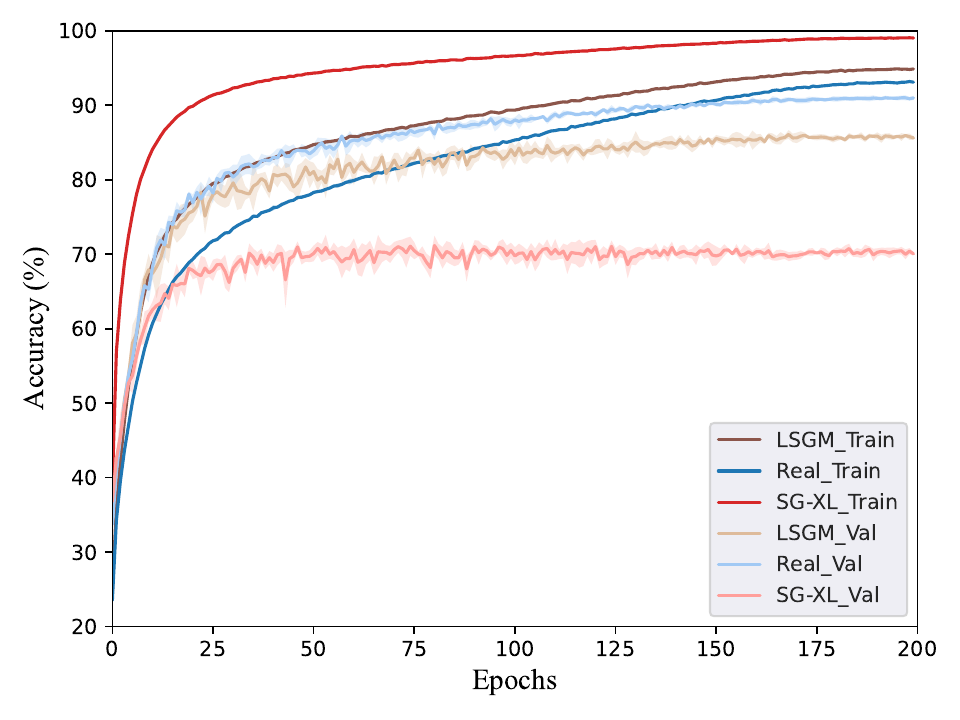}\\
\vspace{-0.3cm}
\caption{The training and validation curves of CIFAR-10 images from different sources over five runs (Random Initialization). Standard deviations are plotted as shaded area. (Zoom in.)}
\label{fig:training-curves-rand}
\end{figure}
}

\newcommand{\figfakelossdistrRand}{
\begin{figure*}[!tbp]
\begin{subfigure}[b]{0.5\textwidth}
\centering
\includegraphics[width=0.8\textwidth]{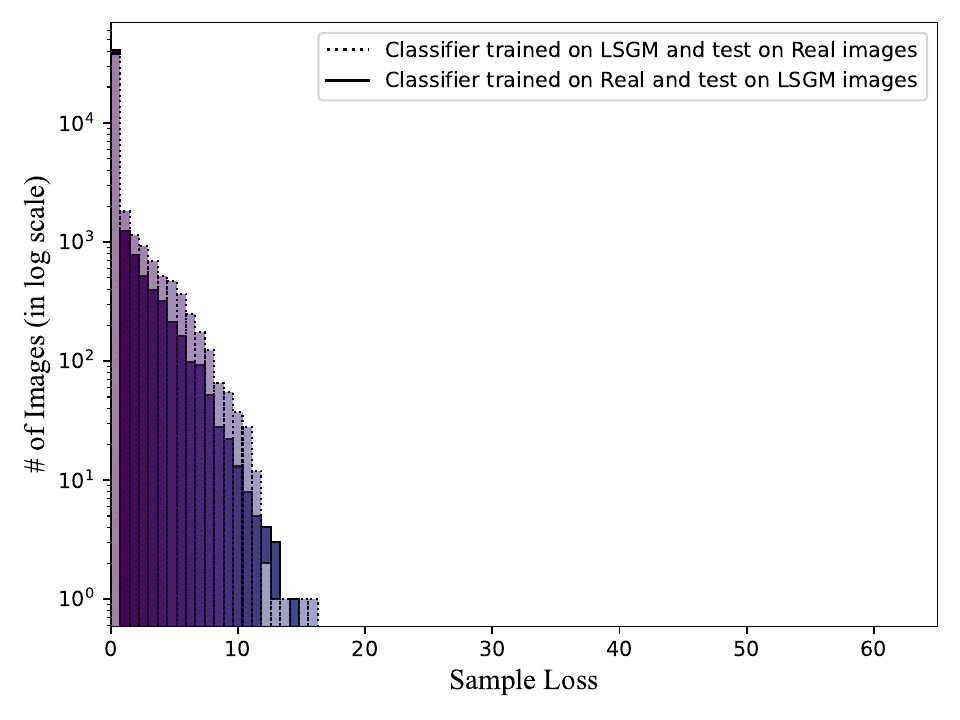}
\vspace{-0.3cm}
\caption{LSGM}
\end{subfigure}
\begin{subfigure}[b]{0.5\textwidth}
\centering
\includegraphics[width=0.8\textwidth]{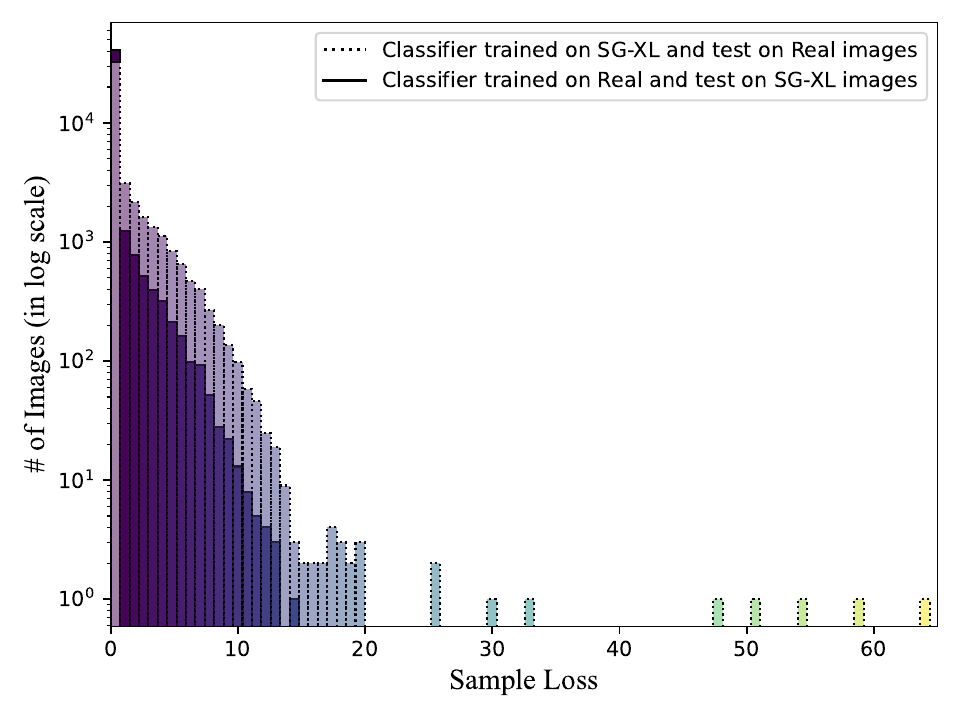}
\vspace{-0.3cm}
\caption{SG-XL}
\end{subfigure}
\vspace{-0.7cm}
\caption{The loss distribution of the samples from different sources.  We overlapped the evaluation of   \emph{Real} $\rightarrow$ \emph{Synthetic} (Solid) and   \emph{Synthetic} $\rightarrow$ \emph{Real} (Dotted) in each subgraph. Note that all models used for evaluation were initialized with random weights.}
\label{fig:fake-loss-distr-rand}
\end{figure*}
}

\newcommand{\figfakelsgm}{
\begin{figure*}[!tbp]
\begin{subfigure}[b]{0.5\textwidth}
\centering
\includegraphics[width=0.9\textwidth]{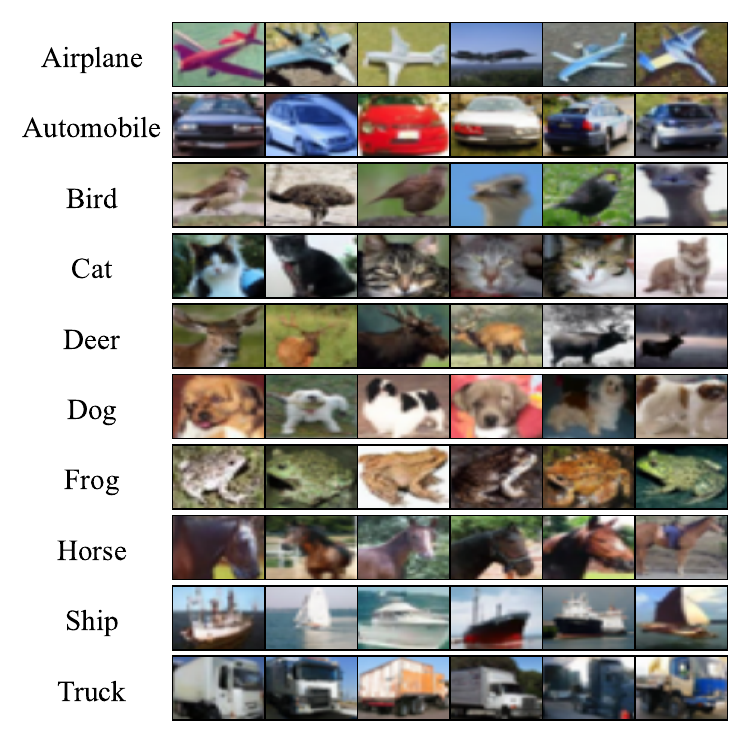}
\vspace{-0.3cm}
\caption{Low-loss Samples}
\end{subfigure}
\begin{subfigure}[b]{0.5\textwidth}
\includegraphics[width=0.9\textwidth]{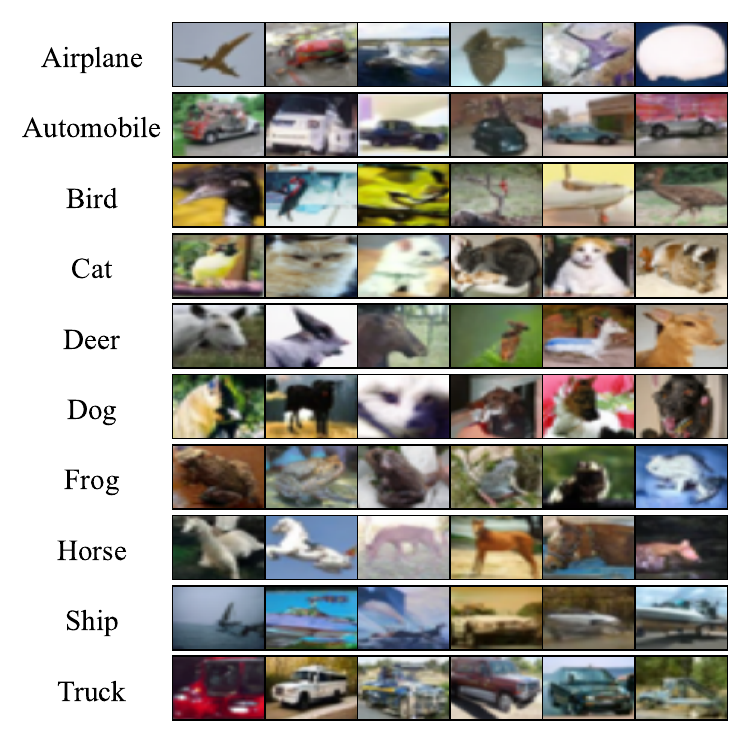}
\vspace{-0.3cm}
\caption{Hihg-loss Samples}
\end{subfigure}
\vspace{-0.7cm}
\caption{The low-loss and high-loss LSGM synthetic samples. Note that the images are in ascending order according to their sample loss. }
\label{fig:fake-lsgm}
\end{figure*}
}

\newcommand{\figfakesgxl}{
\begin{figure*}[!tbp]
\begin{subfigure}[b]{0.5\textwidth}
\centering
\includegraphics[width=0.9\textwidth]{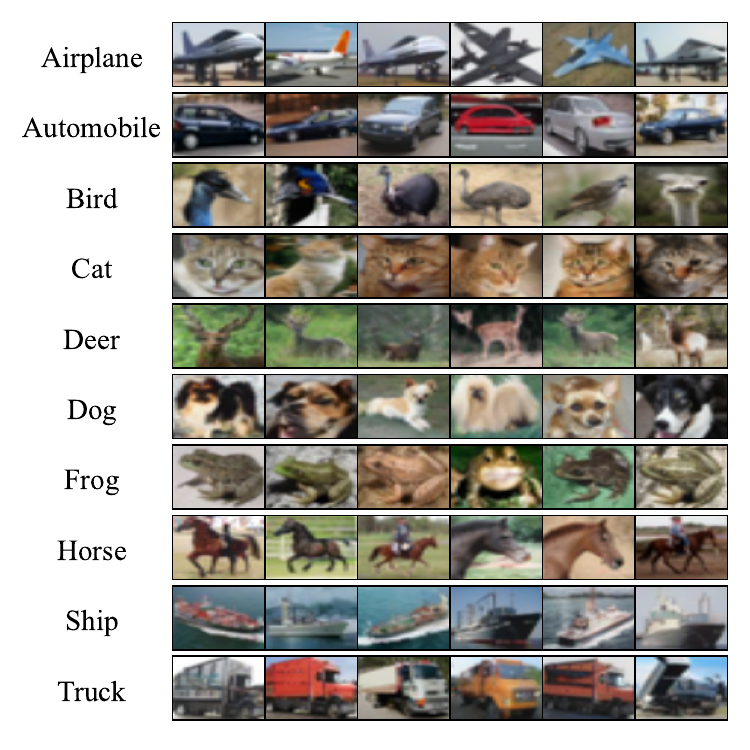}
\vspace{-0.3cm}
\caption{Low-loss Samples}
\end{subfigure}
\begin{subfigure}[b]{0.5\textwidth}
\includegraphics[width=0.9\textwidth]{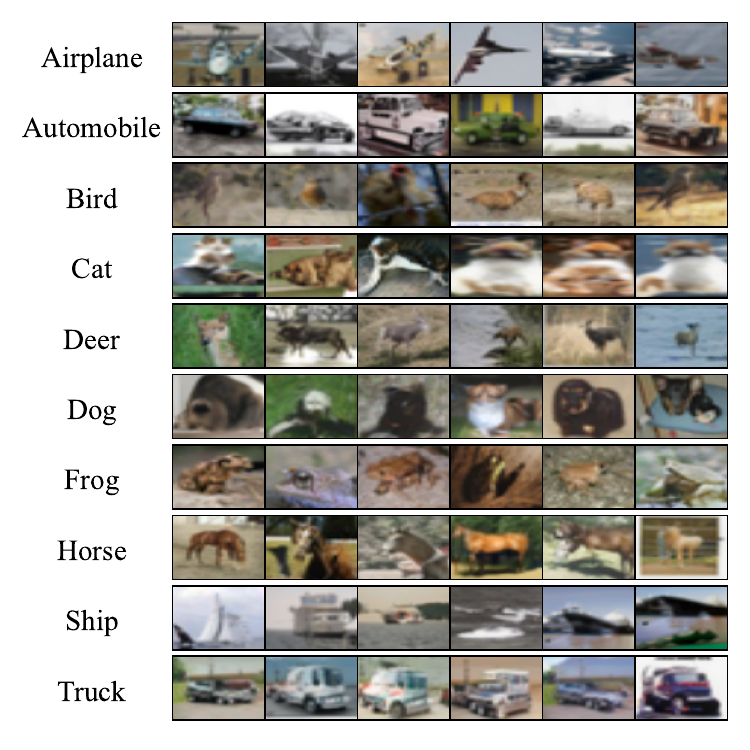}
\vspace{-0.3cm}
\caption{Hihg-loss Samples}
\end{subfigure}
\vspace{-0.7cm}
\caption{The low-loss and high-loss StyleGAN-XL synthetic samples. Note that the images are in ascending order according to their sample loss. }
\label{fig:fake-sgxl}
\end{figure*}
}

\newcommand{\figreallsgm}{
\begin{figure*}[!tbp]
\begin{subfigure}[b]{0.5\textwidth}
\centering
\includegraphics[width=0.9\textwidth]{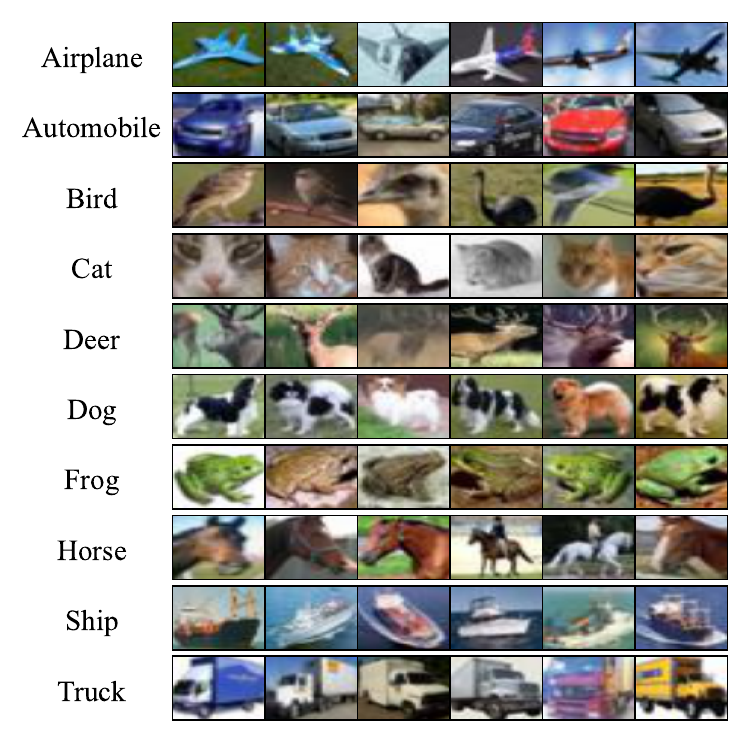}
\vspace{-0.3cm}
\caption{Low-loss Samples}
\end{subfigure}
\begin{subfigure}[b]{0.5\textwidth}
\includegraphics[width=0.9\textwidth]{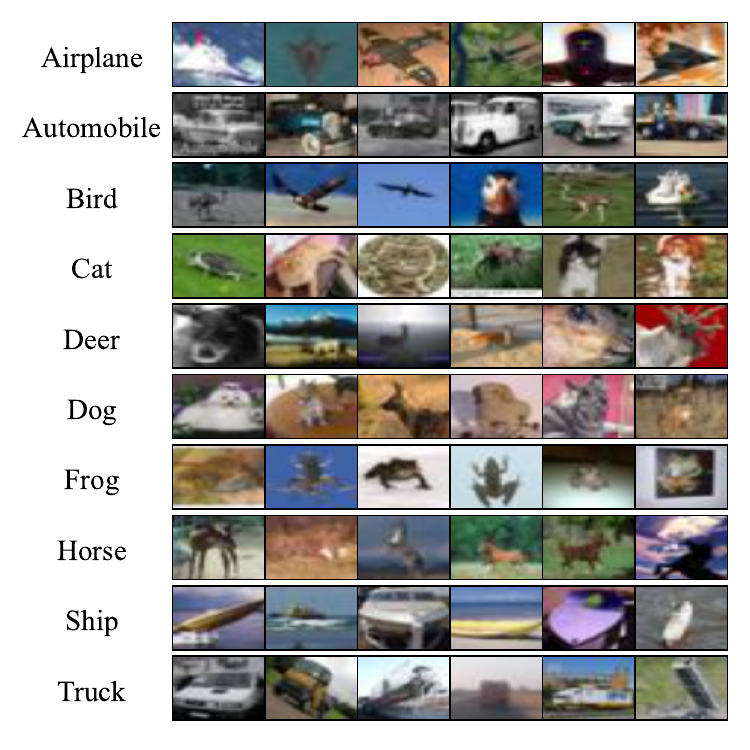}
\vspace{-0.3cm}
\caption{Hihg-loss Samples}
\end{subfigure}
\vspace{-0.7cm}
\caption{The low-loss and high-loss Real samples based on a classifier trained on LSGM synthetic samples. Note that the images are in ascending order according to their sample loss. }
\label{fig:real-lsgm}
\end{figure*}
}

\newcommand{\figrealsgxl}{
\begin{figure*}[!tbp]
\begin{subfigure}[b]{0.5\textwidth}
\centering
\includegraphics[width=0.9\textwidth]{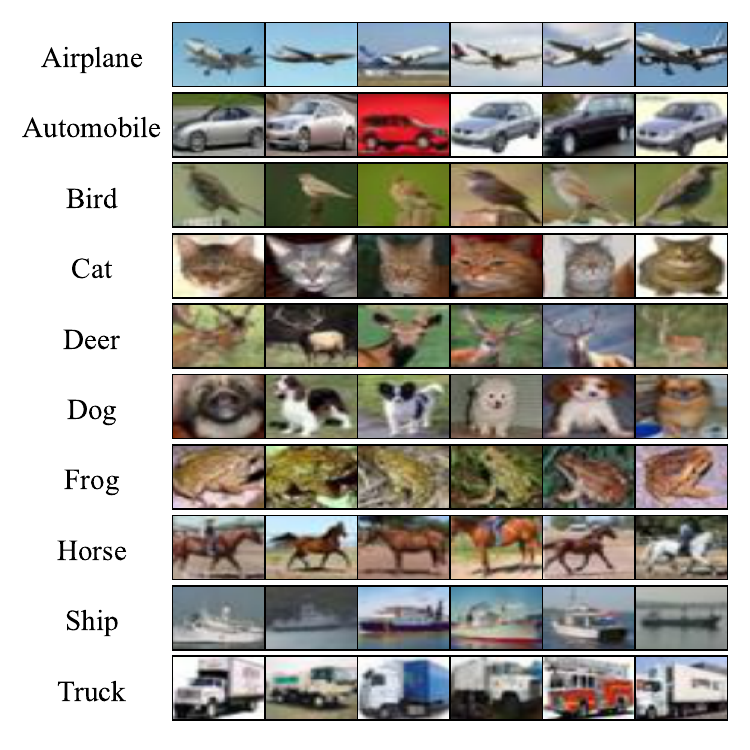}
\vspace{-0.3cm}
\caption{Low-loss Samples}
\end{subfigure}
\begin{subfigure}[b]{0.5\textwidth}
\includegraphics[width=0.9\textwidth]{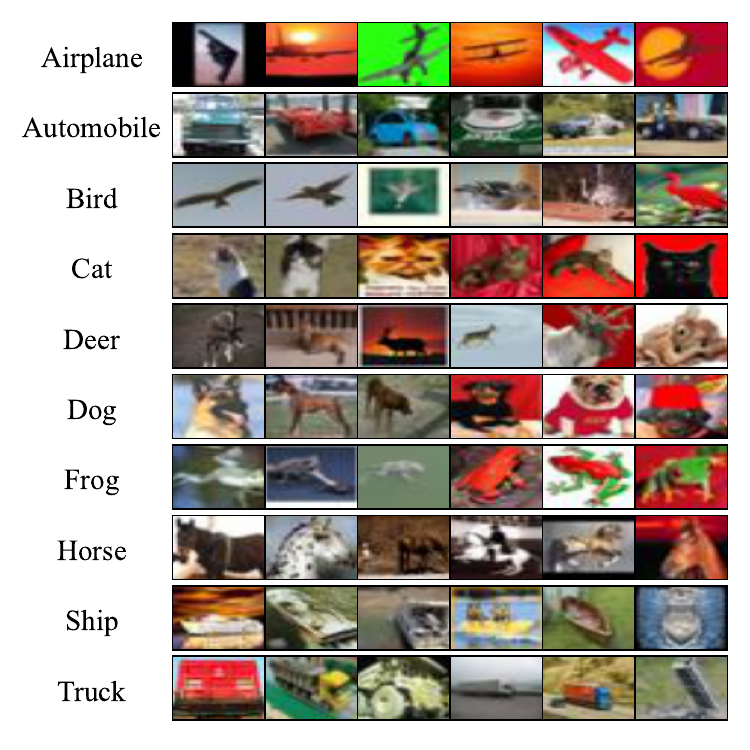}
\vspace{-0.3cm}
\caption{Hihg-loss Samples}
\end{subfigure}
\vspace{-0.7cm}
\caption{The low-loss and high-loss Real samples based on a classifier trained on StyleGAN-XL synthetic samples. Note that the images are in ascending order according to their sample loss. }
\label{fig:real-sgxl}
\end{figure*}
}

\newcommand{\figtraincurvesIMGNET}{
\begin{figure}[!tbp]
\vspace{-0.3cm}
\centering
\includegraphics[width=0.9\linewidth]{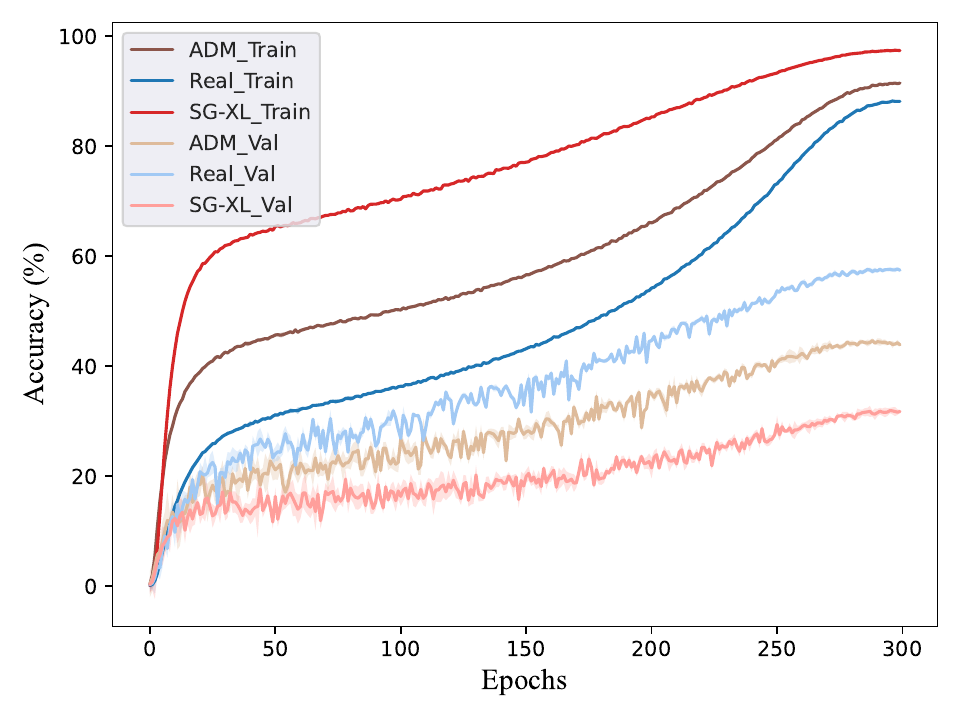}\\
\vspace{-0.3cm}
\caption{The training and validation curves of ImageNet-10\% images from different sources over three runs. Standard deviations are plotted as shaded areas. (Zoom in.)}
\label{fig:training-curves-imgnet}
\vspace{-0.4cm}
\end{figure}
}

\newcommand{\figfakelossdistrRandIMGNET}{
\begin{figure*}[!tbp]
\begin{subfigure}[b]{0.5\textwidth}
\centering
\includegraphics[width=0.8\textwidth]{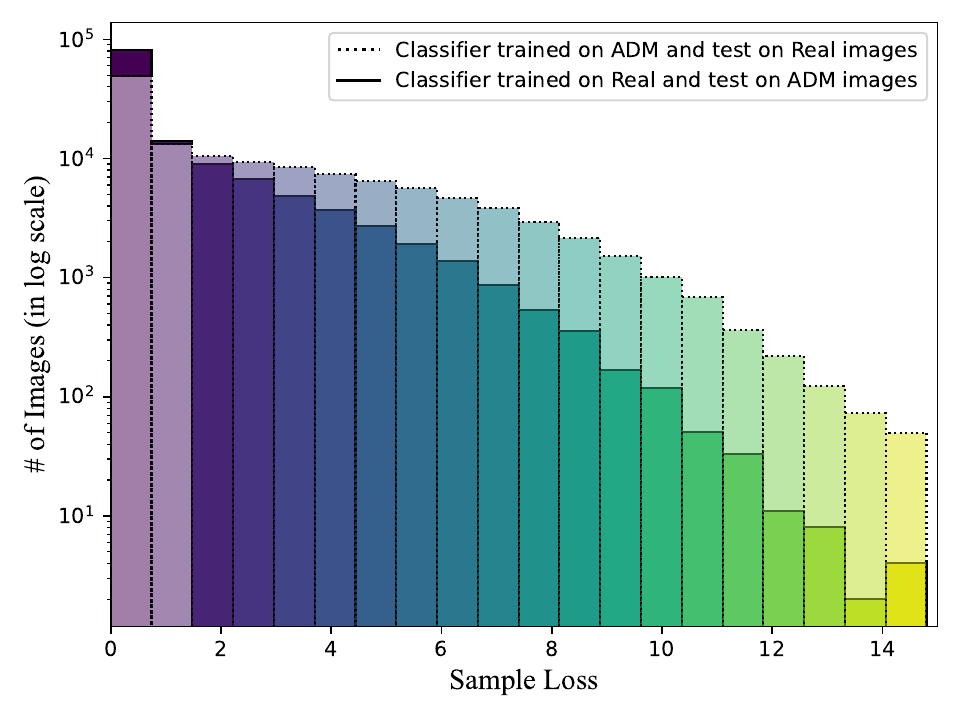}
\vspace{-0.3cm}
\caption{ADM}
\end{subfigure}
\begin{subfigure}[b]{0.5\textwidth}
\centering
\includegraphics[width=0.8\textwidth]{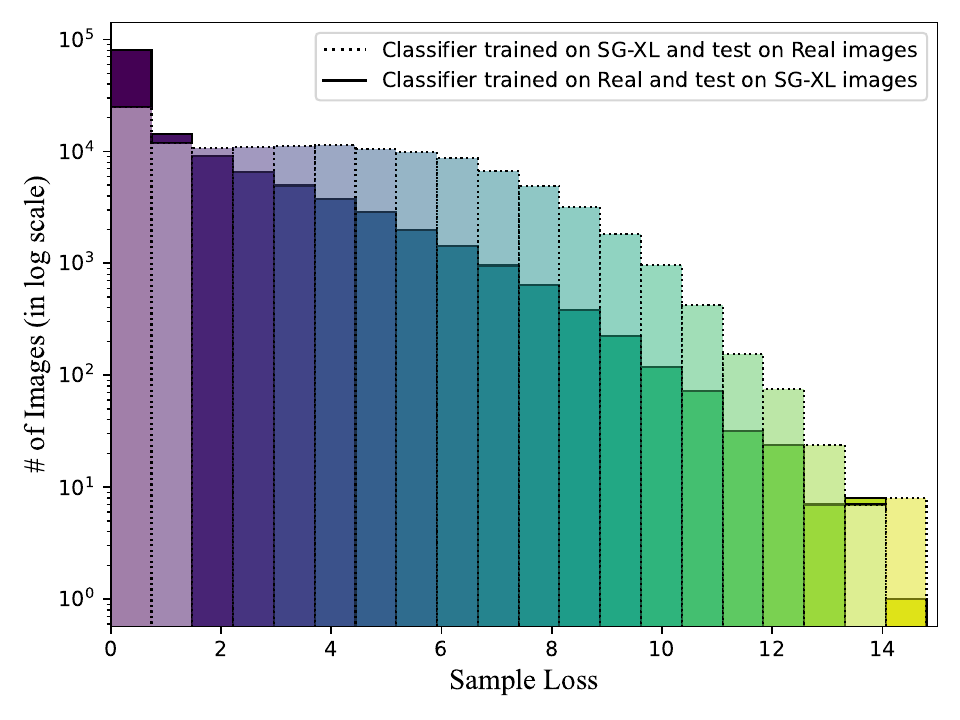}
\vspace{-0.3cm}
\caption{SG-XL}
\end{subfigure}
\vspace{-0.7cm}
\caption{The loss distribution of the samples from different sources.  We overlapped the evaluation of   \emph{Real} $\rightarrow$ \emph{Synthetic} (Solid) and   \emph{Synthetic} $\rightarrow$ \emph{Real} (Dotted) in each subgraph. Note that all models used for evaluation were initialized with random weights.}
\label{fig:fake-loss-distr-rand-imgnet}
\end{figure*}
}

\newcommand{\figfakesgIMGNET}{
\begin{figure*}[!tbp]
\begin{subfigure}[b]{1.0\textwidth}
\centering
\includegraphics[width=1.0\textwidth]{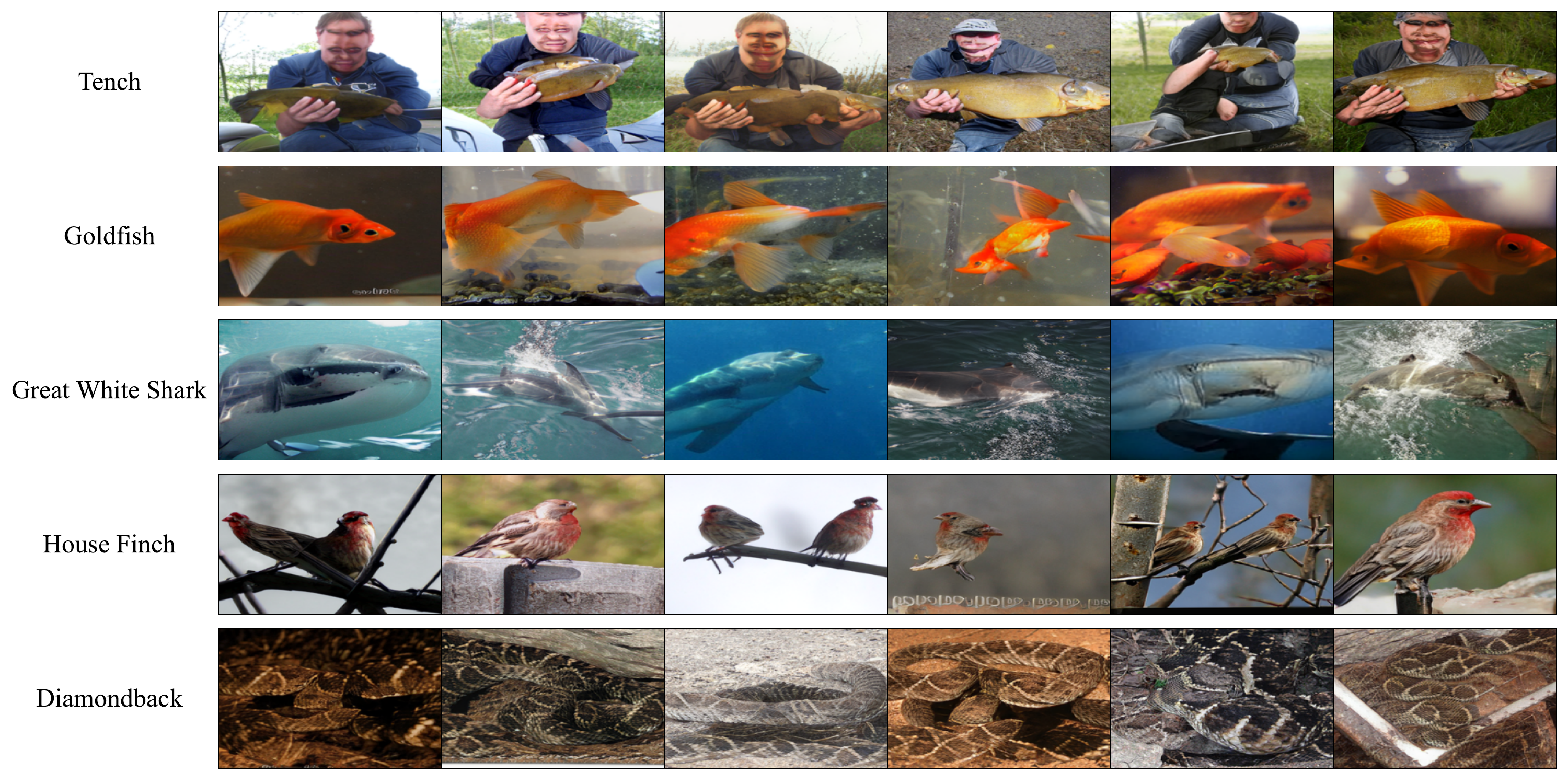}
\vspace{-0.6cm}
\caption{Low-loss Samples}
\end{subfigure}
\begin{subfigure}[b]{1.0\textwidth}
\includegraphics[width=1.0\textwidth]{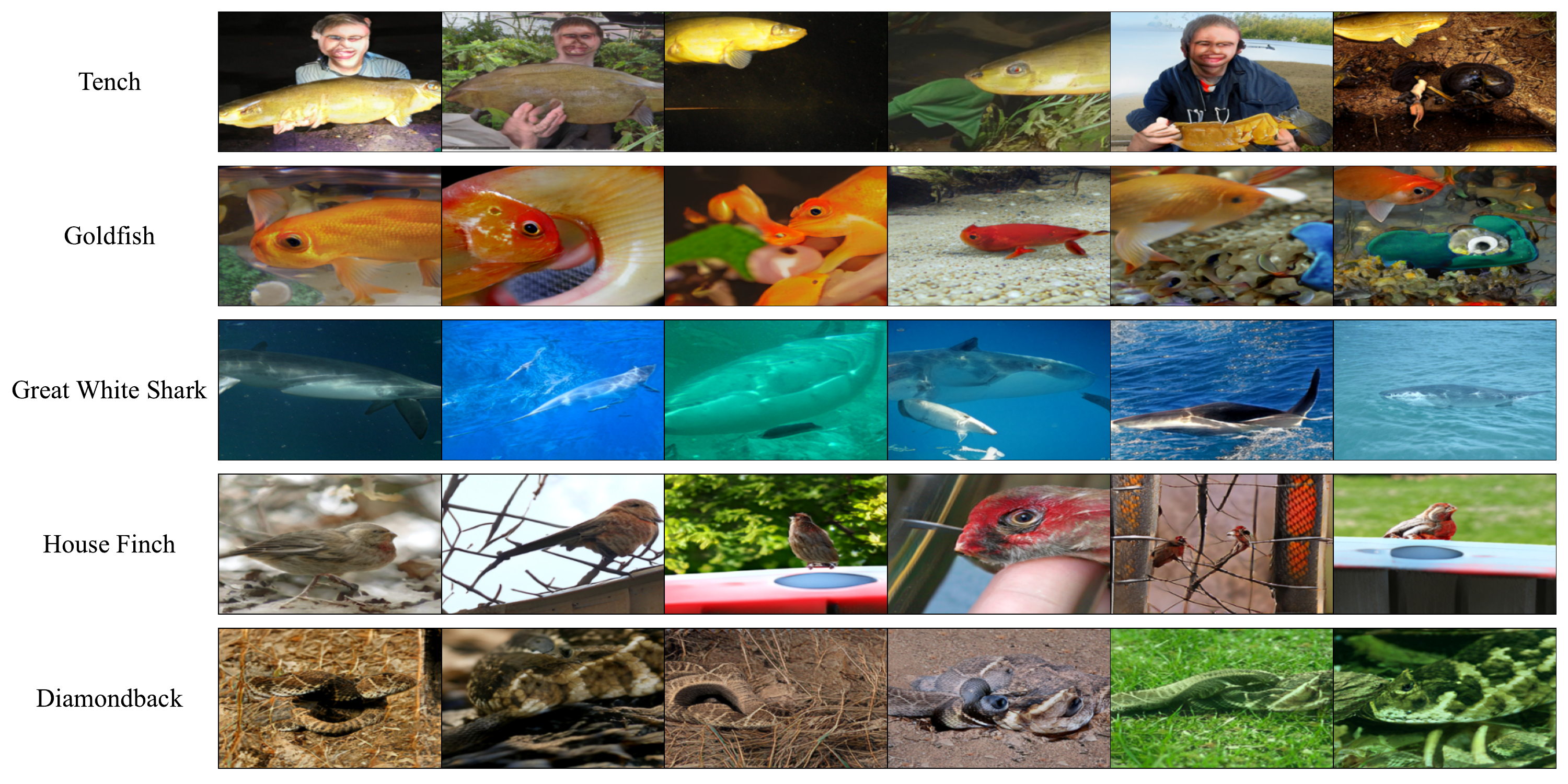}
\vspace{-0.6cm}
\caption{Hihg-loss Samples}
\end{subfigure}
\vspace{-0.7cm}
\caption{The low-loss and high-loss StyleGAN-XL synthetic samples. Note that the images are in ascending order according to their sample loss. }
\label{fig:fake-sg-imgnet}
\end{figure*}
}

\newcommand{\figfakegdIMGNET}{
\begin{figure*}[!tbp]
\begin{subfigure}[b]{1.0\textwidth}
\centering
\includegraphics[width=1.0\textwidth]{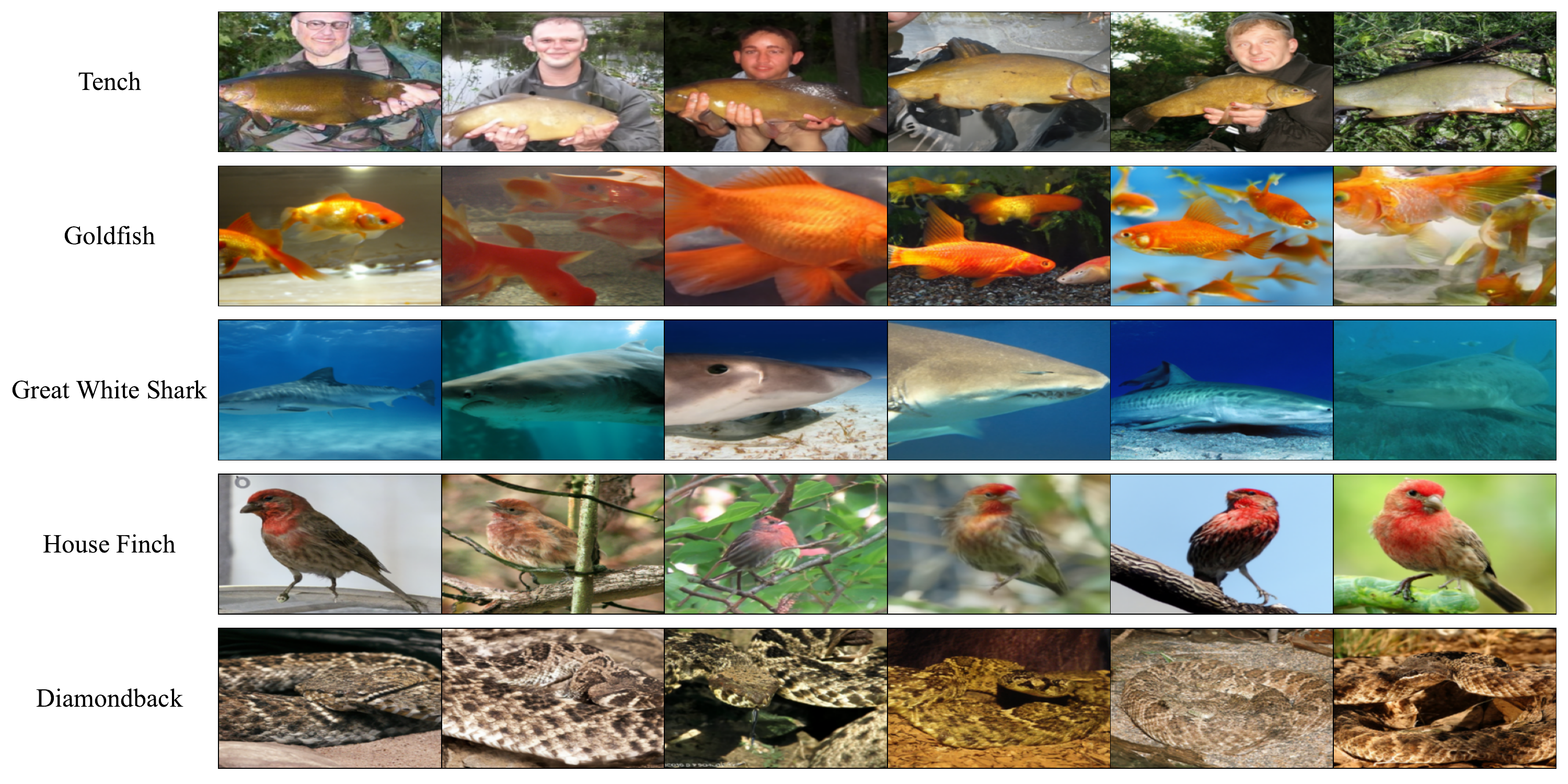}
\vspace{-0.6cm}
\caption{Low-loss Samples}
\end{subfigure}
\begin{subfigure}[b]{1.0\textwidth}
\includegraphics[width=1.0\textwidth]{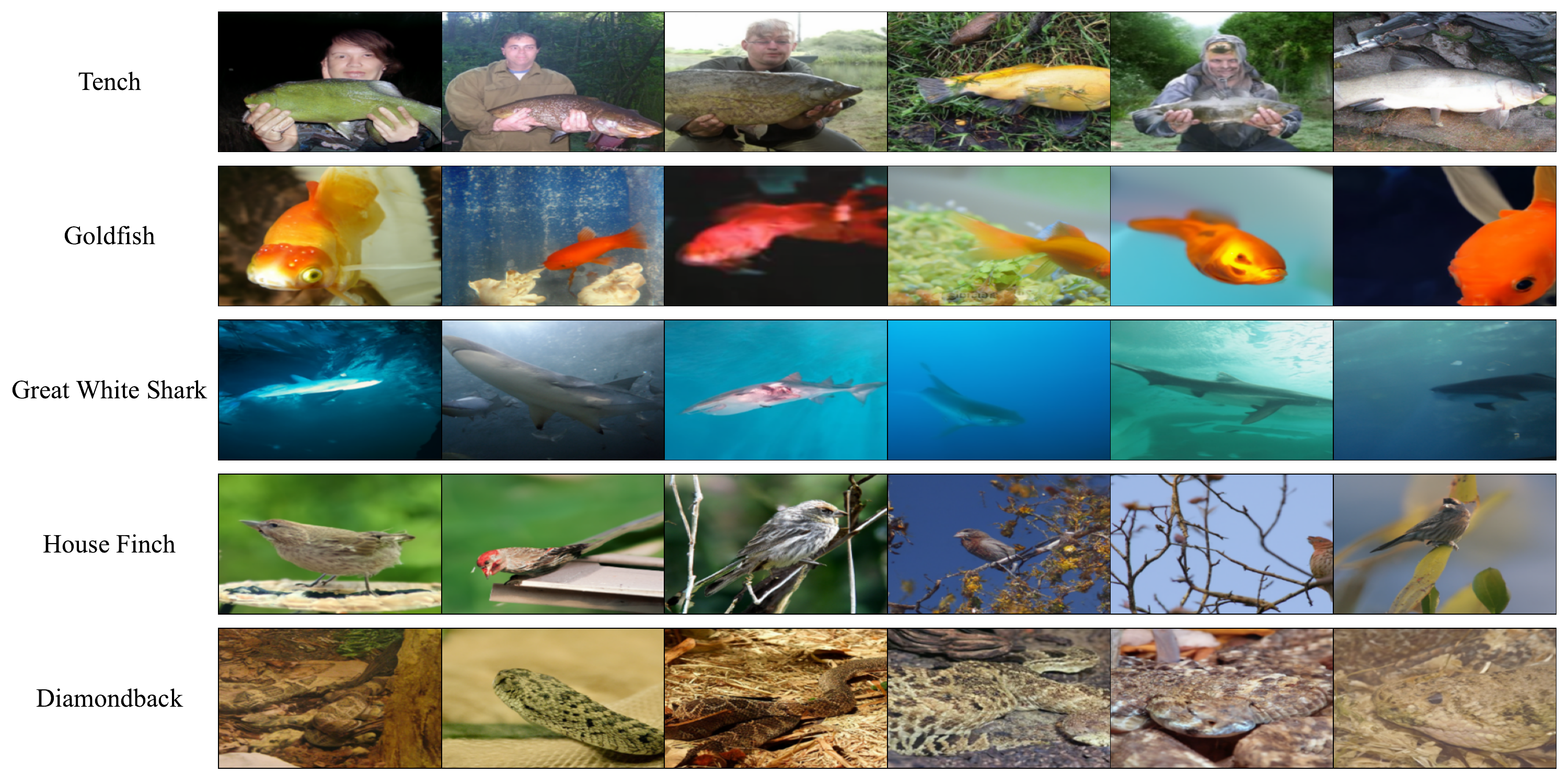}
\vspace{-0.6cm}
\caption{Hihg-loss Samples}
\end{subfigure}
\vspace{-0.7cm}
\caption{The low-loss and high-loss ADM synthetic samples. Note that the images are in ascending order according to their sample loss. }
\label{fig:fake-gd-imgnet}
\end{figure*}
}

\newcommand{\figrealsgIMGNET}{
\begin{figure*}[!tbp]
\begin{subfigure}[b]{1.0\textwidth}
\centering
\includegraphics[width=1.0\textwidth]{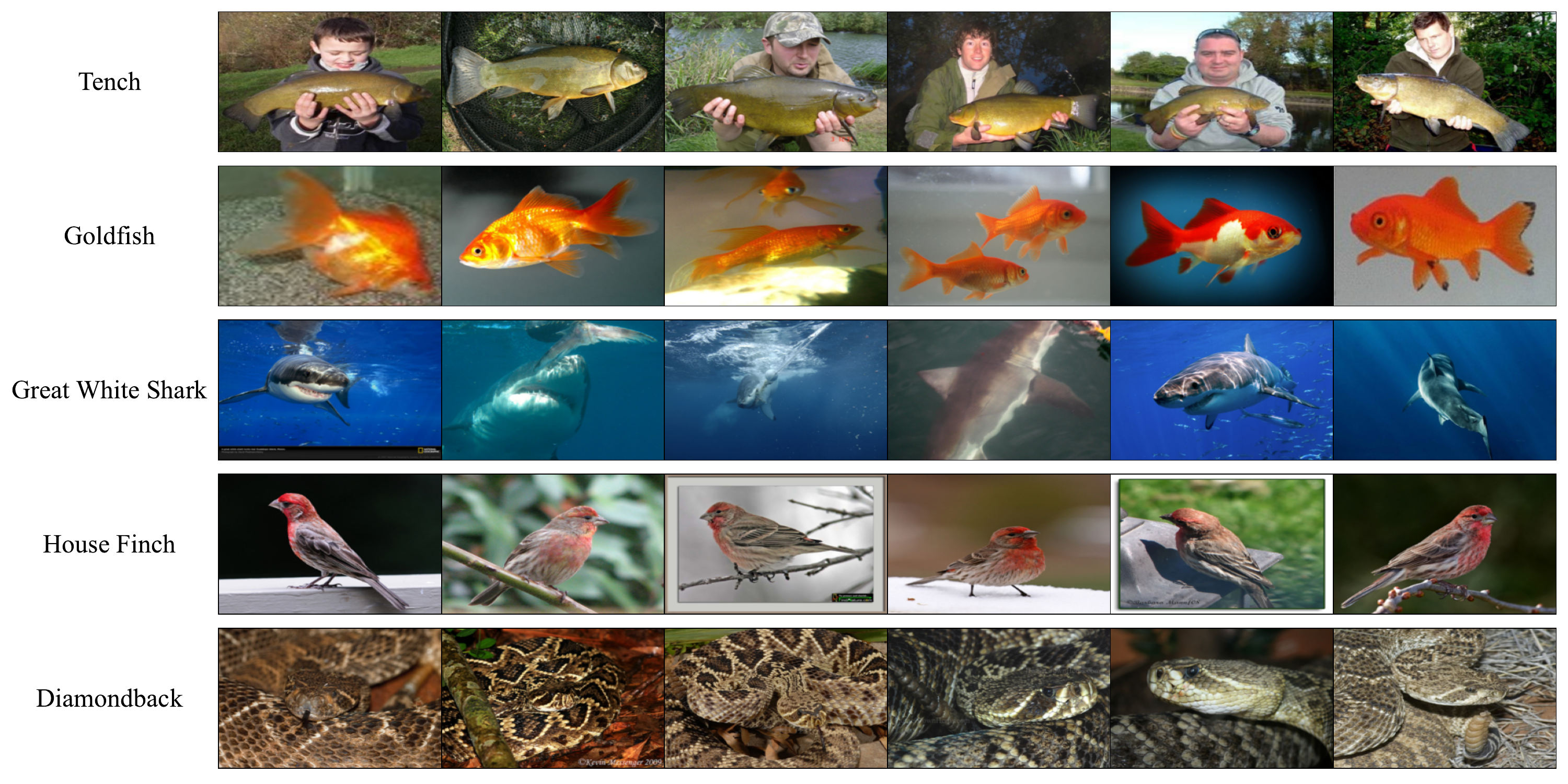}
\vspace{-0.6cm}
\caption{Low-loss Samples}
\end{subfigure}
\begin{subfigure}[b]{1.0\textwidth}
\includegraphics[width=1.0\textwidth]{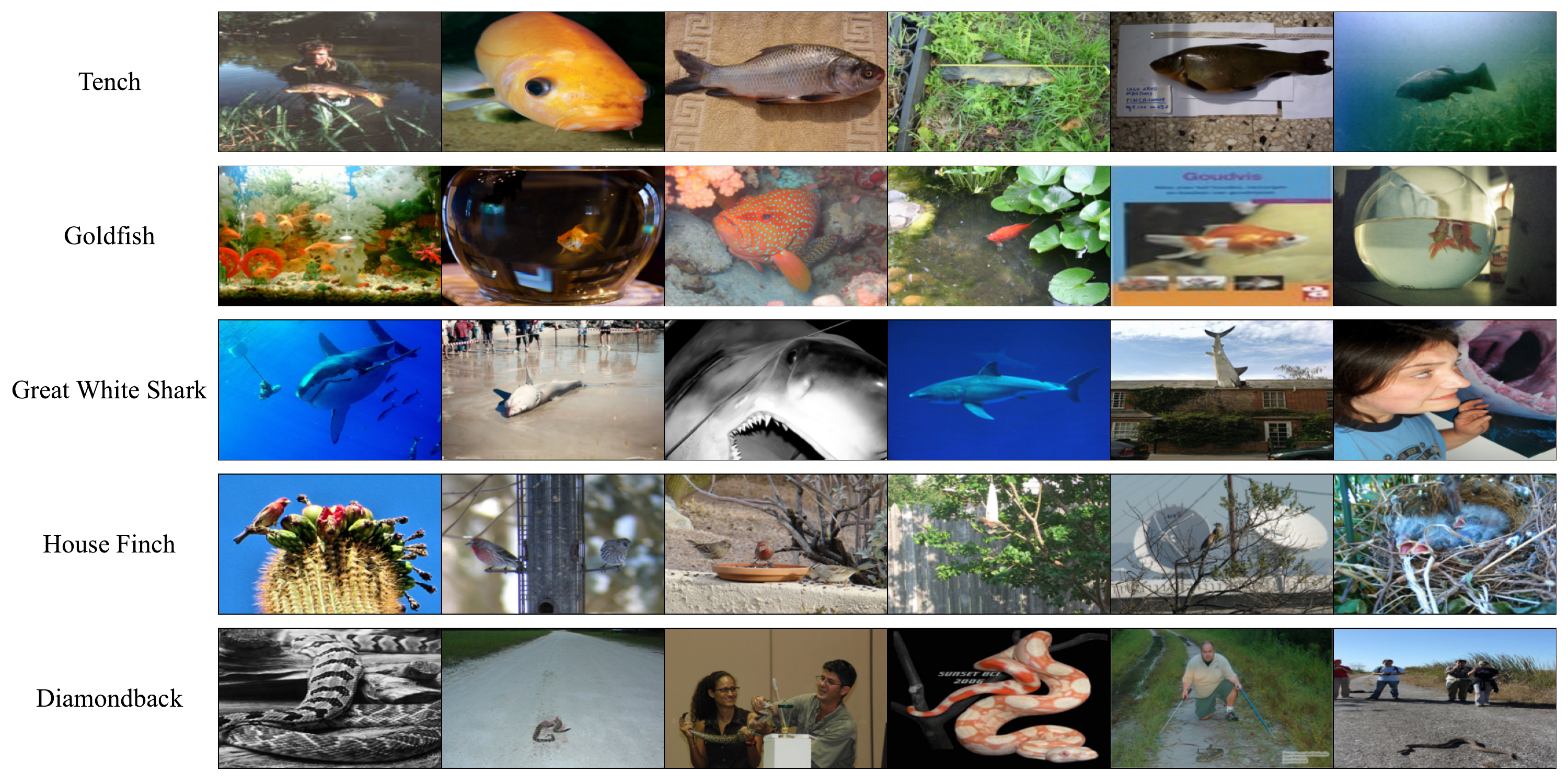}
\vspace{-0.6cm}
\caption{Hihg-loss Samples}
\end{subfigure}
\vspace{-0.7cm}
\caption{The low-loss and high-loss Real samples based on a classifier trained on StyleGAN-XL synthetic samples. Note that the images are in ascending order according to their sample loss. }
\label{fig:real-sg-imgnet}
\end{figure*}
}

\newcommand{\figrealgdIMGNET}{
\begin{figure*}[!tbp]
\begin{subfigure}[b]{1.0\textwidth}
\centering
\includegraphics[width=1.0\textwidth]{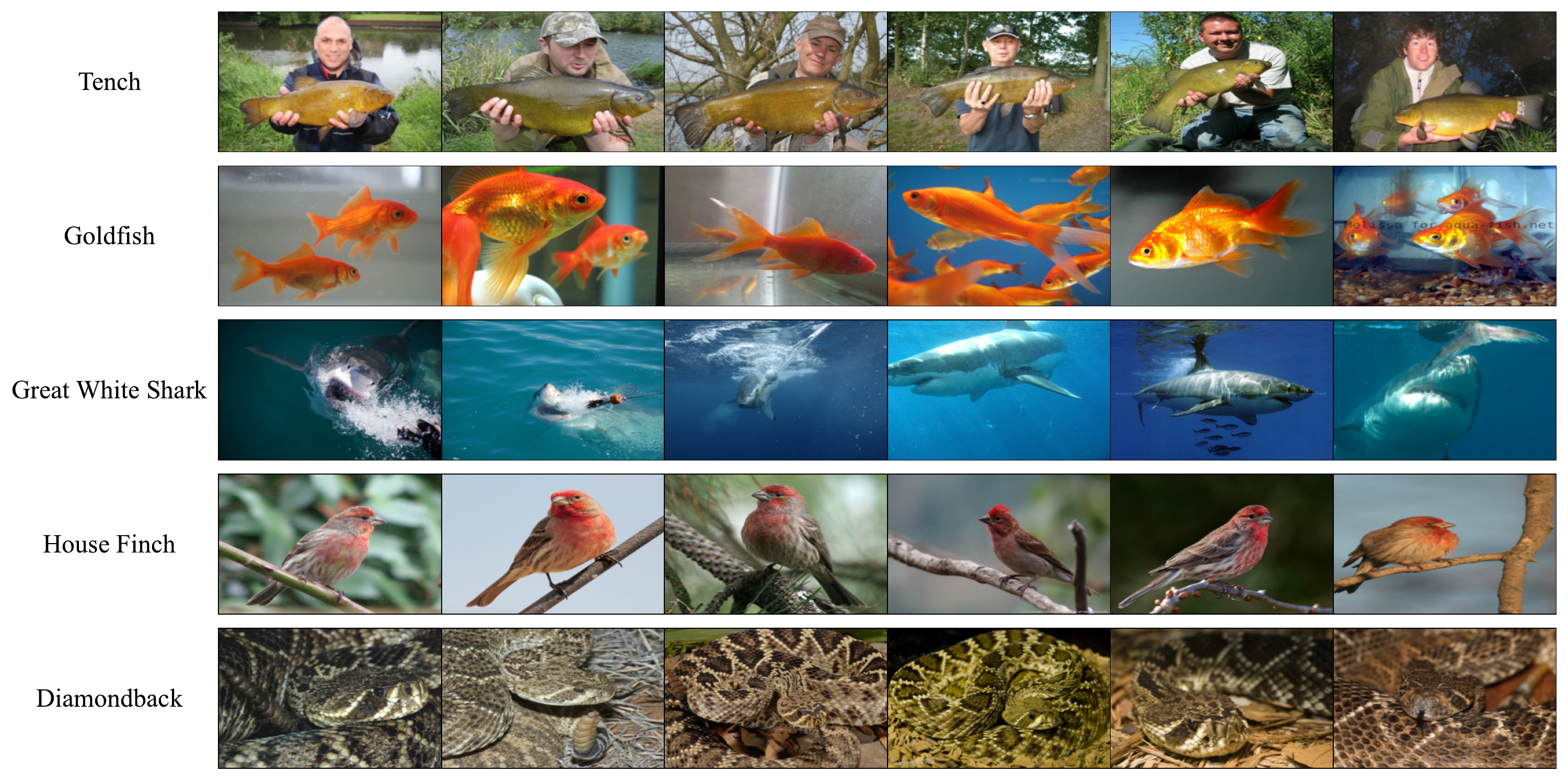}
\vspace{-0.6cm}
\caption{Low-loss Samples}
\end{subfigure}
\begin{subfigure}[b]{1.0\textwidth}
\includegraphics[width=1.0\textwidth]{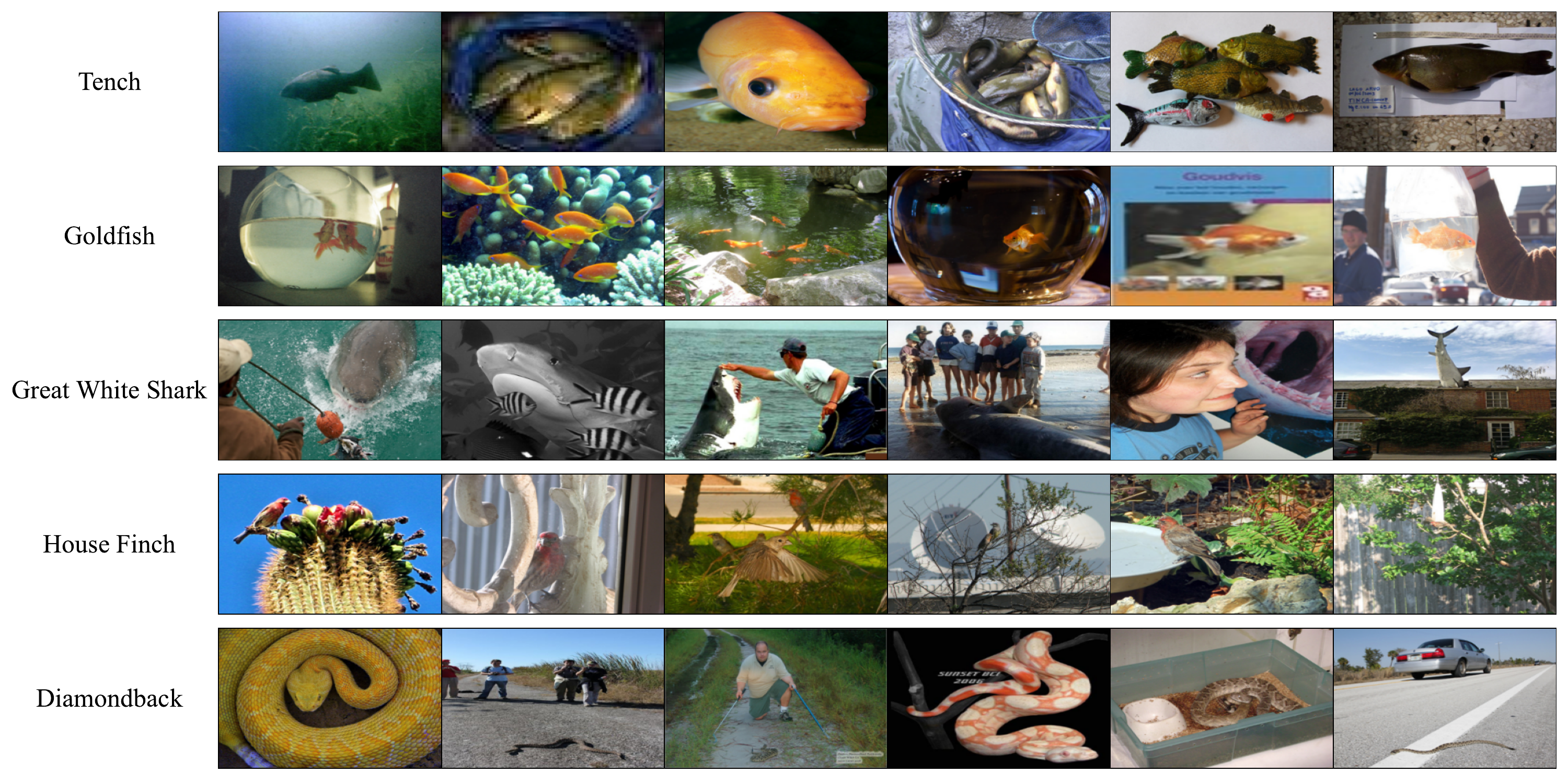}
\vspace{-0.6cm}
\caption{Hihg-loss Samples}
\end{subfigure}
\vspace{-0.7cm}
\caption{The low-loss and high-loss Real samples based on a classifier trained on ADM synthetic samples. Note that the images are in ascending order according to their sample loss. }
\label{fig:real-gd-imgnet}
\end{figure*}
}


%% file: tables.tex
\newcommand{\tabnonmutualdg}{
\begin{table}[!tbp]
\caption{The achieved accuracy of classifiers trained and tested on different sources of images. Note that the results were all acquired from the training sets, therefore the high accuracy in the diagonal line indicates that the classifiers have converged on their own training set.}
\vspace{-2.2mm}
\label{tab:non_mutual_dg}
\centering
\newcolumntype{x}{>{\centering\arraybackslash\hspace{0pt}}p{10.5mm}}
\footnotesize{
\begin{tabular}{l@{\hspace{1.5mm}}l@{\hspace{0mm}} x@{\hspace{-0.1mm}} x@{\hspace{-0.1mm}} x@{\hspace{-0.1mm}}  x }
\toprule
& \textbf{Source of Test Images} &Real &LSGM & SG-XL \\
\midrule \midrule 
&\textbf{Classifier Trained on Real} &96.34 &93.54 &98.68 \\
&\textbf{Classifier Trained on LSGM Images} &85.76 &97.71 &99.11 \\
&\textbf{Classifier Trained on SG-XL Images} &78.90 &91.14 &99.82 
\\
\bottomrule
\end{tabular}
}
\vspace{-1.3mm}
\end{table}
}

\newcommand{\tabfakesubsets}{
\begin{table}[!tbp]
\caption{The achieved accuracy of classifiers trained on the datasets augmented by different subsets of synthetic images. We note the size of the training sets in brackets. }
\vspace{-2.2mm}
\label{tab:fake_subsets}
\centering
\newcolumntype{x}{>{\centering\arraybackslash\hspace{0pt}}p{20mm}}
\footnotesize{
\begin{tabular}{l@{\hspace{1.5mm}}l@{\hspace{0mm}} x@{\hspace{-0.1mm}} x@{\hspace{-0.1mm}} x@{\hspace{-0.1mm}}  x }
\toprule
&  &LSGM &SG-XL \\
\midrule 
\midrule 
&\textbf{Real+Small Losses (67.5k)}  &91.99 (+0.19) &91.91 (+0.11)\\
&\textbf{Real+Large Losses (67.5k)}  &92.50 (+0.70) &91.97 (+0.17) \\
\midrule
& \textbf{Real Only (45k)} &\multicolumn{2}{c}{91.80}
\\
\bottomrule
\end{tabular}
}
\vspace{-1.2mm}
\end{table}
}

\newcommand{\tabotherfakes}{
\begin{table*}[!tbp]
\caption{The achieved accuracy of classifiers trained on different synthetic datasets. (A) The networks were trained from scratch. (B) The networks were initialized by pretrained ImageNet weights.  Note that for a fair comparison,  we omit the second stage of ASG \cite{chen2020automated}, which uses reinforcement learning for selective parameter update.}
\vspace{-2.2mm}
\label{tab:other_fake_cls}
\centering
\newcolumntype{x}{>{\centering\arraybackslash\hspace{0pt}}p{15mm}}
\footnotesize{
\begin{tabular}{l@{\hspace{1.5mm}}l@{\hspace{0mm}} x@{\hspace{-0.1mm}} x@{\hspace{-0.1mm}} x@{\hspace{-0.1mm}}  x@{\hspace{-0.1mm}} x@{\hspace{-0.1mm}} x@{\hspace{-0.1mm}} x@{\hspace{-0.1mm}} x@{\hspace{-0.1mm}} x@{\hspace{-0.1mm}} x@{\hspace{-0.1mm}}  x }
\toprule
 & &\textbf{Dataset} & \multicolumn{2}{c}{CIFAR-10} & \multicolumn{2}{c}{CUB} &\multicolumn{2}{c}{Flower} &SDI-A &SDI-B &SDI-C  \\
 & &\textbf{Size} & \multicolumn{2}{c}{45,000} & \multicolumn{2}{c}{4,521} &\multicolumn{2}{c}{1,010} &8,000 &8,000 &8,000  \\
\cmidrule(lr){4-5}\cmidrule(lr){6-7}\cmidrule(lr){8-9} \cmidrule(lr){10-12}
& &\textbf{Source} &LSGM &SG-XL &Proj-GAN &SG-XL &Proj-GAN &SG-XL &\multicolumn{3}{c}{DT-GAN}  \\
\midrule \midrule 
&\multirow{6}{3.5em}{\textbf{(A)}} &\textbf{Baseline} &85.08 &72.22 &12.30 &10.80 &13.54 &13.07 &64.72 &86.80 &61.14 \\
& &\textbf{ASG \cite{chen2020automated}}&80.66 &71.68 &1.64 &1.83 &5.32 &5.12 &32.78 &69.40 &31.43  \\
& &\textbf{CSG \cite{chen2021contrastive}}&82.47 &72.94 &1.41 &1.86 &8.21 &8.04 &43.94 &74.55 &32.19  \\
& &\textbf{Ours (L1)}&\textbf{85.58} &75.07 &11.93 &\textbf{13.94} &13.89 &13.69 &64.73 &87.20 &\textbf{64.00} \\
& &\textbf{Ours (KL)}&84.76 &\textbf{77.45} &\textbf{17.11} &13.45 &\textbf{17.16}  &\textbf{15.53} &\textbf{66.36} &\textbf{88.40} &62.52 \\
\cmidrule(lr){3-12}
& &\textbf{Real} & \multicolumn{2}{c}{90.27} & \multicolumn{2}{c}{18.82} &\multicolumn{2}{c}{35.59} &83.20 &87.8 &72.38  \\
\midrule \midrule 
&\multirow{6}{3.5em}{\textbf{(B)}} &\textbf{Baseline} &85.57 &78.94 &33.15 &27.20 &32.40 &28.42 &53.64 &74.00 &52.95 \\
& &\textbf{ASG \cite{chen2020automated}}&82.40 &77.86 &\textbf{43.62} &\textbf{39.90} &5.63 &5.47 &\textbf{67.78} &\textbf{87.95} &50.47  \\
& &\textbf{CSG \cite{chen2021contrastive}}&82.67 &79.01 &\textbf{43.62} &38.81 &5.45 &4.33 &62.33 &84.45 &50.86  \\
& &\textbf{Ours (L1)}&\textbf{85.74} &79.87 &32.89 &27.32 &30.91 &26.00 &51.63 &73.40 &53.52 \\
& &\textbf{Ours (KL)}&84.73 &\textbf{80.66} &34.50 &28.93 &\textbf{33.52}  &\textbf{30.01} &57.64 &85.20 &\textbf{53.90} \\
\cmidrule(lr){3-12}
& &\textbf{Real} & \multicolumn{2}{c}{91.80} & \multicolumn{2}{c}{43.15} &\multicolumn{2}{c}{52.12} &66.36 &90.00 &63.81  \\
\bottomrule
\end{tabular}
}
\vspace{-1.2mm}
\end{table*}
}

\newcommand{\tabrealshots}{
\begin{table*}[!tbp]
\caption{The achieved accuracy of classifiers trained with different methods under various setting: \textbf{RG} indicate Real Guidance, \textbf{PG-F} means only apply Pretrained Guidance to the synthetic data, and \textbf{PG-R} means only apply Pretrained Guidance to the real data. Note that we denote the baseline two-stage domain adaption as \textbf{Adp.} and baseline data augmentation as \textbf{Aug.}.}
\vspace{-2.2mm}
\label{tab:real_shots}
\centering
\newcolumntype{x}{>{\centering\arraybackslash\hspace{0pt}}p{20mm}}
\footnotesize{
\begin{tabular}{l@{\hspace{1.5mm}}l@{\hspace{-5mm}} x@{\hspace{-13mm}} x@{\hspace{-12mm}} x@{\hspace{-10mm}} x@{\hspace{-8mm}} x@{\hspace{-6mm}} x@{\hspace{-8mm}} x@{\hspace{-6mm}} x@{\hspace{-8mm}} x@{\hspace{-4mm}} x@{\hspace{-8mm}} x@{\hspace{-8mm}} x@{\hspace{-4mm}}  x }
\toprule
&\textbf{Dataset} & & & &\multicolumn{2}{c}{\quad\! CIFAR-10 (10-shots)} & \multicolumn{2}{c}{\quad\quad CUB} &\multicolumn{2}{c}{\quad\! Flower} &SDI-A &SDI-B &SDI-C  \\
&\textbf{Syn-to-Real Ratio} & & & &\multicolumn{2}{c}{\quad\! 450:1} & \multicolumn{2}{c}{\quad\quad 1:1} &\multicolumn{2}{c}{\quad\! 1:1} &2:1 &2:1 &2:1  \\
&\textbf{Source}&\textbf{RG} &\textbf{PG-R} &\textbf{PG-F} &LSGM &SG-XL &Proj-GAN &SG-XL &Proj-GAN &SG-XL &\multicolumn{3}{c}{\quad\quad DT-GAN}    \\
\midrule \midrule 
&\textbf{Baseline (Real)} &- &- &-  & \multicolumn{2}{c}{\quad\! 91.80} & \multicolumn{2}{c}{\quad\quad 43.15} &\multicolumn{2}{c}{\quad\! 52.12} &66.36 &90.00 &63.81   \\
&\textbf{Baseline (Fake)} &- &- &- &85.57 &78.94 &33.15 &27.20 &32.40 &28.42 &53.64 &74.00 &52.95  \\
&\textbf{Baseline (Adp.)} &- &- &- &84.81	&78.38 &39.56	&37.94	&37.71	&36.70	&81.09	&91.40	&83.43
   \\
&\textbf{Baseline (Aug.)} &- &- &- &85.71 &79.38 &52.94 &52.28 &60.54 &60.61 &86.73 &90.80 &86.86   \\
\midrule

&\textbf{ADDA\cite{Tzeng2017AdversarialDD}} &- &- &- &78.73 &74.42 &23.47	&22.07	&29.30	&27.79	&56.72	&70.00	&57.33
 \\
&\textbf{DADA\cite{dada}} &- &- &- &83.39	&76.56	&26.42	&23.68	&20.59	&18.93	&55.09	&74.20	&53.33
 \\
\midrule
&\textbf{DANN\cite{ganin2015unsupervised}} &- &- &- &86.52 &81.01 &57.34	&55.79	&59.88	&58.80	&77.09	&87.60	&77.14
  \\
&\textbf{LTDA\cite{JamalLongtail_DA}} &- &- &- &81.67 &73.89 &24.72	&18.92	&20.22	&17.39	&66.00	&83.20	&74.29
 \\
&\textbf{A-GEM\cite{Chaudhry2018EfficientLL}} &- &- &- &\textbf{86.91} &80.33 &33.38	&29.37	&33.09	&31.88	&66.91	&91.60	&67.81
 \\

\midrule
&\multirow{5}{3.5em}{\textbf{Ours}} &- &v &v  &86.25	&80.96	&54.71	&53.51	&61.47	&61.10	&89.27	&93.00	&87.61
 \\
& &v &- &-  &\textbf{86.91} &80.33 &59.39	&57.87	&64.57	&63.31	&\textbf{90.73}	&\textbf{96.20}	&\textbf{91.43}
 \\
& &v &v &-  &85.59 &81.25 &\textbf{60.65}	&58.69	&66.04	&64.88	&85.27	&95.20	&90.48
 \\
& &v &- &v  &86.58 &79.70 &59.90	&58.51	&65.00	&64.74	&89.82	&95.60	&88.57
 \\
& &v &v &v  &85.94 &\textbf{81.93} &60.17	&\textbf{59.11}	&\textbf{66.24}	&\textbf{65.05}
 &87.64 &95.80 &89.33 \\

\bottomrule
\end{tabular}
}
\vspace{-1.2mm}
\end{table*}
}

\newcommand{\tabdatasetinfo}{
\begin{table}[!tbp]
\caption{The quantitative results of the sampled synthetic datasets from commonly used evaluation metrics: FID, Precision/Recall \cite{Kynkaanniemi2019}, and Diversity/Coverage \cite{ferjad2020icml}. Note that the reported FIDs were computed between the original dataset and the sampled synthetic dataset, therefore might differ from the reported FID of the checkpoints used for sampling.}
\vspace{-2.2mm}
\label{tab:datasetinfo}
\centering
\newcolumntype{x}{>{\centering\arraybackslash\hspace{0pt}}p{12mm}}
\footnotesize{
\begin{tabular}{l@{\hspace{0mm}}l@{\hspace{0mm}} x@{\hspace{-1.9mm}} x@{\hspace{-0.1mm}} x@{\hspace{-0.1mm}} x@{\hspace{-0.1mm}} x@{\hspace{-0.1mm}} x@{\hspace{-0.1mm}} x }
\toprule
&\textbf{Dataset} &\textbf{Method}  &\textbf{FID}	&\textbf{Precision}	&\textbf{Recall}	&\textbf{Diversity}	&\textbf{Coverage}
   \\
\midrule \midrule 
&\multirow{2}{3.5em}{\textbf{CIFAR-10}}&{\cite{vahdat2021score}}  &16.55 &0.66 &0.65 &0.70 &0.74 \\
&&{\cite{Sauer2021ARXIV}} &32.97 &0.75 &0.34 &1.01 &0.64\\
 \midrule 
&\multirow{2}{3.5em}{\textbf{CUB}}&{\cite{Sauer2021NEURIPS}}  &7.64 &0.80  &0.59 &1.16&0.92 \\
&&{\cite{Sauer2021ARXIV}}  &10.05 &0.88 &0.39 &1.52&0.92 \\
 \midrule 
&\multirow{2}{3.5em}{\textbf{Flower}}&{\cite{Sauer2021NEURIPS}}  &26.91 &0.83  &0.69 &0.97&0.93 \\
&&{\cite{Sauer2021ARXIV}}  &29.83 &0.90 &0.48 &1.18&0.91 \\
 \midrule 
&\textbf{SDI-A} &\multirow{3}{1.6em}{\cite{Wang_2022_BMVC}}  &131.24 &0.04  &0.01 &0.01 &0.01 \\
&\textbf{SDI-B} & &140.39  &0.30 &0.01 &0.14 &0.05 \\
&\textbf{SDI-C} &  &116.16 &0.01 &0.01 &0.01 &0.01 \\
\bottomrule
\end{tabular}
}
\vspace{-1.2mm}
\end{table}
}

\newcommand{\tabcifarsmall}{
\begin{table}[!tbp]
\caption{The achieved accuracy of classifiers trained on synthetic CIFAR-10 datasets, where a simplified version of ResNet-50 was used for training images at resolution 32. }
\vspace{-2.2mm}
\label{tab:cifar_small}
\centering
\newcolumntype{x}{>{\centering\arraybackslash\hspace{0pt}}p{20mm}}
\footnotesize{
\begin{tabular}{l@{\hspace{1.5mm}}l@{\hspace{0mm}} x@{\hspace{-0.1mm}} x@{\hspace{-0.1mm}}   x }
\toprule

&\textbf{Method}  &LSGM  &SG-XL   \\
\midrule \midrule 
&\textbf{Baseline}  &78.55 &70.68 \\
&\textbf{ASG \cite{chen2020automated}} &76.63 &66.31  \\
&\textbf{Ours (L1)}  &78.33 &70.80   \\
&\textbf{Ours (KL)}  &\textbf{80.11} &\textbf{74.21} \\
\midrule 
&\textbf{Real}  &\multicolumn{2}{c}{84.53} \\
\bottomrule
\end{tabular}
}
\vspace{-1.2mm}
\end{table}
}

\newcommand{\tabnonmutualdgRand}{
\begin{table}[!tbp]
\caption{The achieved accuracy of classifiers (Random Initialization) trained and tested on different sources of images. Note that the results were all acquired from the training sets, therefore the high accuracy in the diagonal line indicates that the classifiers have converged on their own training set.}
\vspace{-2.2mm}
\label{tab:non_mutual_dg_rand}
\centering
\newcolumntype{x}{>{\centering\arraybackslash\hspace{0pt}}p{10.5mm}}
\footnotesize{
\begin{tabular}{l@{\hspace{1.5mm}}l@{\hspace{0mm}} x@{\hspace{-0.1mm}} x@{\hspace{-0.1mm}} x@{\hspace{-0.1mm}}  x }
\toprule
& \textbf{Source of Test Images} &Real &LSGM & SG-XL \\
\midrule \midrule 
&\textbf{Classifier Trained on Real} &99.85 &90.09 &97.25 \\
&\textbf{Classifier Trained on LSGM Images} &85.82 &99.83 &98.18 \\
&\textbf{Classifier Trained on SG-XL Images} &72.32 &79.43 &99.86 
\\
\bottomrule
\end{tabular}
}
\vspace{-1.2mm}
\end{table}
}

\newcommand{\tabfakesubsetsRand}{
\begin{table}[!tbp]
\caption{The achieved accuracy of classifier (Random Initialization) trained on the datasets augmented by different subsets of synthetic images. We note the size of the training sets in brackets. }
\vspace{-2.2mm}
\label{tab:fake_subsets_rand}
\centering
\newcolumntype{x}{>{\centering\arraybackslash\hspace{0pt}}p{20mm}}
\footnotesize{
\begin{tabular}{l@{\hspace{1.5mm}}l@{\hspace{0mm}} x@{\hspace{-0.1mm}} x@{\hspace{-0.1mm}} x@{\hspace{-0.1mm}}  x }
\toprule
&  &LSGM &SG-XL \\
\midrule 
\midrule 
&\textbf{Real+Small Losses (67.5k)}  &91.54 (+1.27) &90.97 (+0.70)\\
&\textbf{Real+Large Losses (67.5k)}  &92.50 (+2.23) &91.38 (+1.11) \\
\midrule
& \textbf{Real Only (45k)} &\multicolumn{2}{c}{90.27}
\\
\bottomrule
\end{tabular}
}
\vspace{-1.2mm}
\end{table}
}

\newcommand{\tabotherfakesHP}{
\begin{table*}[!tbp]
\caption{The intensity of Pretrained Guidance ($\lambda_{3}$) used in \autoref{tab:other_fake_cls}. (A) The networks were train from scratch. (B) The networks were initialized by pretrained ImageNet weights.}
\vspace{-2.2mm}
\label{tab:other_fake_cls_hp}
\centering
\newcolumntype{x}{>{\centering\arraybackslash\hspace{0pt}}p{15mm}}
\footnotesize{
\begin{tabular}{l@{\hspace{1.5mm}}l@{\hspace{0mm}} x@{\hspace{-0.1mm}} x@{\hspace{-0.1mm}} x@{\hspace{-0.1mm}}  x@{\hspace{-0.1mm}} x@{\hspace{-0.1mm}} x@{\hspace{-0.1mm}} x@{\hspace{-0.1mm}} x@{\hspace{-0.1mm}} x@{\hspace{-0.1mm}} x@{\hspace{-0.1mm}}  x }
\toprule
 & &\textbf{Dataset} & \multicolumn{2}{c}{CIFAR-10} & \multicolumn{2}{c}{CUB} &\multicolumn{2}{c}{Flower} &SDI-A &SDI-B &SDI-C  \\
\cmidrule(lr){4-5}\cmidrule(lr){6-7}\cmidrule(lr){8-9} \cmidrule(lr){10-12}
& &\textbf{Source} &LSGM &SG-XL &Proj-GAN &SG-XL &Proj-GAN &SG-XL &\multicolumn{3}{c}{DT-GAN}  \\
\midrule \midrule 
&\multirow{2}{3.5em}{\textbf{(A)}} &\textbf{Ours(L1)}&1 &1 &1 &5 &1 &5 &1 &5 &5 \\
& &\textbf{Ours(KL)}&1,000 &1,000 &600 &600 &600  &600 &75 &1 &1 \\
\midrule \midrule 
&\multirow{2}{3.5em}{\textbf{(B)}} &\textbf{Ours (L1)}&1 &1 &1 &1 &1 &1 &1 &5 &10 \\
& &\textbf{Ours (KL)}&1,000 &1,000 &45 &50 &45  &1 &1 &1,000 &1 \\
\bottomrule
\end{tabular}
}
\vspace{-1.2mm}
\end{table*}
}

\newcommand{\tabrealshotsHP}{
\begin{table*}[!tbp]
\caption{The intensity of Real Guidance ($\lambda_{1}$ and $\lambda_{2}$) and Pretrained Guidance ($\lambda_{2}$) used in \autoref{tab:real_shots} (ImageNet Initialization). We report the selected $\lambda_{1}$, $\lambda_{2}$ and $\lambda_{3}$ in each entry.}
\vspace{-2.2mm}
\label{tab:real_shots_hp}
\centering
\newcolumntype{x}{>{\centering\arraybackslash\hspace{0pt}}p{15mm}}
\footnotesize{
\begin{tabular}{l@{\hspace{1.5mm}}l@{\hspace{0mm}}   x@{\hspace{-0.1mm}} x@{\hspace{-0.1mm}} x@{\hspace{-0.1mm}} x@{\hspace{-0.1mm}} x@{\hspace{-0.1mm}} x@{\hspace{-0.1mm}} x@{\hspace{-0.1mm}} x@{\hspace{-0.1mm}} x@{\hspace{-0.1mm}}  x }
\toprule
&\textbf{Dataset}  &\multicolumn{2}{c}{\quad\! CIFAR-10 (10-shots)} & \multicolumn{2}{c}{\quad\quad CUB} &\multicolumn{2}{c}{\quad\! Flower} &SDI-A &SDI-B &SDI-C  \\
&\textbf{Syn-to-Real Ratio} &\multicolumn{2}{c}{\quad\! 450:1} & \multicolumn{2}{c}{\quad\quad 1:1} &\multicolumn{2}{c}{\quad\! 1:1} &2:1 &2:1 &2:1  \\
&\textbf{Source} &LSGM &SG-XL &Proj-GAN &SG-XL &Proj-GAN &SG-XL &\multicolumn{3}{c}{\quad\quad DT-GAN}    \\
\midrule \midrule 

&{\textbf{$\lambda_{1}$}}   &0	&0	&1	&1	&1	&1	&1	&1	&1
 \\
&{\textbf{$\lambda_{2}$}}   &1	&1	&2	&2	&2	&2	&1	&1	&1
 \\
&{\textbf{$\lambda_{3}$}}   &100	&100	&50	&50	&50	&50	&1.75	&1.75	&0.1
 \\

\bottomrule
\end{tabular}
}
\vspace{-1.2mm}
\end{table*}
}

\newcommand{\tabrandrealshots}{
\begin{table*}[!tbp]
\caption{The achieved accuracy of classifiers (Random Initialization) trained with different methods under various setting: \textbf{RG} indicate Real Guidance, \textbf{PG-F} means only apply Pretrained Guidance to the synthetic data, and \textbf{PG-R} means only apply Pretrained Guidance to the real data.   Note that we denote the baseline two-stage domain adaption as \textbf{Adp.} and baseline data augmentation as \textbf{Aug.}.}
\vspace{-2.2mm}
\label{tab:real_shots_rand}
\centering
\newcolumntype{x}{>{\centering\arraybackslash\hspace{0pt}}p{20mm}}
\footnotesize{
\begin{tabular}{l@{\hspace{1.5mm}}l@{\hspace{-5mm}} x@{\hspace{-13mm}} x@{\hspace{-12mm}} x@{\hspace{-10mm}} x@{\hspace{-8mm}} x@{\hspace{-6mm}} x@{\hspace{-8mm}} x@{\hspace{-6mm}} x@{\hspace{-8mm}} x@{\hspace{-4mm}} x@{\hspace{-8mm}} x@{\hspace{-8mm}} x@{\hspace{-4mm}}  x }
\toprule
&\textbf{Dataset} & & & &\multicolumn{2}{c}{\quad\! CIFAR-10 (10-shots)} & \multicolumn{2}{c}{\quad\quad CUB} &\multicolumn{2}{c}{\quad\! Flower} &SDI-A &SDI-B &SDI-C  \\
&\textbf{Syn-to-Real Ratio} & & & &\multicolumn{2}{c}{\quad\! 450:1} & \multicolumn{2}{c}{\quad\quad 1:1} &\multicolumn{2}{c}{\quad\! 1:1} &2:1 &2:1 &2:1  \\
&\textbf{Source}&\textbf{RG} &\textbf{PG-R} &\textbf{PG-F} &LSGM &SG-XL &Proj-GAN &SG-XL &Proj-GAN &SG-XL &\multicolumn{3}{c}{\quad\quad DT-GAN}    \\
\midrule \midrule 
&\textbf{Baseline (Real)} &- &- &-  & \multicolumn{2}{c}{\quad\! 90.27} & \multicolumn{2}{c}{\quad\quad 18.82} &\multicolumn{2}{c}{\quad\! 35.59} &83.20 &87.80 &72.38   \\
&\textbf{Baseline (Fake)} &- &- &- &85.08 &72.20 &12.30 &10.80 &13.54 &13.07 &64.72 &86.60 &61.14  \\
&\textbf{Baseline (Adp.)} &- &- &- &84.95	&71.93	&21.86	&21.76	&20.11	&19.25	&73.45	&85.60	&64.76
   \\
&\textbf{Baseline (Aug.)} &- &- &- &85.10	&72.13	&33.78	&32.44	&36.84	&37.93	&85.82	&89.40	&\textbf{87.05}
   \\
\midrule
&\textbf{ADDA \cite{Tzeng2017AdversarialDD}} &- &- &- &83.87	&68.83	&10.87	&11.21	&13.84	&13.97	&54.00	&73.20	&52.38
 \\
&\textbf{DADA \cite{dada}} &- &- &- &83.01	&71.83	&12.80	&12.09	&13.72	&14.16	&53.45	&81.00	&51.24
 \\
\midrule
&\textbf{DANN \cite{ganin2015unsupervised}} &- &- &- &85.23	&73.47	&31.62	&31.76	&35.78	&36.60	&80.18	&91.40	&56.95
  \\
&\textbf{LTDA \cite{JamalLongtail_DA}} &- &- &- &82.13	&74.47	&21.89	&15.77	&16.03	&13.38	&70.91	&88.40	&71.43
 \\
&\textbf{A-GEM \cite{Chaudhry2018EfficientLL}} &- &- &- &\textbf{86.37}	&72.79	&12.72	&12.66	&14.76	&14.73	&74.55	&88.40	&71.62
 \\
\midrule
&\multirow{5}{3.5em}{\textbf{Ours}} &- &v &v  &84.92	&77.66	&\textbf{41.59}	&\textbf{39.49}	&\textbf{42.52}	&\textbf{41.27}	&\textbf{89.27}	&93.00	&86.10
 \\
& &v &- &-  &\textbf{86.37}	&72.79	&33.04	&31.22	&36.64	&36.47	&88.18	&\textbf{93.20}	&77.71
 \\
& &v &v &-  &85.59	&78.70	&36.27	&33.46	&35.94	&36.29	&78.91	&91.80	&73.90
 \\
& &v &- &v  &86.15	&73.19	&36.53	&35.72	&38.59	&38.83	&88.91	&92.60	&81.52
 \\
& &v &v &v  &86.00	&\textbf{78.88}	&39.60	&37.95	&40.14	&38.77	&86.18	&92.60	&76.38
 \\

\bottomrule
\end{tabular}
}
\vspace{-1.2mm}
\end{table*}
}

\newcommand{\tabdgmtraining}{
\begin{table}[!tbp]
\caption{The training and sampling statistic of various deep generative models on the CUB-Bird and Oxford-Flower dataset. For training, we denote the converged model by ``v", otherwise ``x". For sampling, we report the number of classes that met the sampling requirements. }
\vspace{-2.2mm}
\label{tab:dgm_training}
\centering
\newcolumntype{x}{>{\centering\arraybackslash\hspace{0pt}}p{14mm}}
\footnotesize{
\begin{tabular}{l@{\hspace{1.5mm}}l@{\hspace{0mm}} x@{\hspace{-0.1mm}} x@{\hspace{-0.1mm}} x@{\hspace{-0.1mm}} x@{\hspace{-0.1mm}}  x@{\hspace{-0.1mm}}  x }
\toprule

 &\textbf{Dataset} &\textbf{Stage}  &\textbf{LSGM}  &\textbf{Fast-GAN} &\textbf{Proj-GAN} &\textbf{SG-XL}   \\
\midrule \midrule 
&\multirow{2}{3.5em}{\textbf{CUB}} &Training  &x &v &v &v \\
& &Sampling &0 / 200 &173 / 200 &199 / 200 & 194 / 200  \\
\midrule
&\multirow{2}{3.5em}{\textbf{Flower}} &Training  &v &v &v &v \\
& &Sampling &69 / 102 &67 / 102 &102 / 102 & 101 / 102  \\
\bottomrule
\end{tabular}
}
\vspace{-1.2mm}
\end{table}
}

\newcommand{\tabtrainvaltestIMGNET}{
\begin{table}[!tbp]
\caption{The achieved accuracy of classifiers trained and tested on different sources of images. Note that the validation and test sets are from the official ImageNet validation set, containing only real images. The training sets are the one used to train the respective classifier.}
\vspace{-2.2mm}
\label{tab:train_val_test_imgnet}
\centering
\newcolumntype{x}{>{\centering\arraybackslash\hspace{0pt}}p{10.5mm}}
\footnotesize{
\begin{tabular}{l@{\hspace{1.5mm}}l@{\hspace{0mm}} x@{\hspace{-0.1mm}} x@{\hspace{-0.1mm}} x@{\hspace{-0.1mm}}  x }
\toprule
& \textbf{Source of Test Images} &Train &Val & Test \\
\midrule \midrule 
&\textbf{Classifier Trained on Real} &99.78 &57.67 &57.67 \\
&\textbf{Classifier Trained on ADM Images} &99.69 &44.67 &43.79 \\
&\textbf{Classifier Trained on SG-XL Images} &99.99 &31.94 &31.55
\\
\bottomrule
\end{tabular}
}
\vspace{-1.3mm}
\end{table}
}

\newcommand{\tabnonmutualdgIMGNET}{
\begin{table}[!tbp]
\caption{The achieved accuracy of classifiers trained and tested on different sources of images. Note that the results were all acquired from the training sets, therefore the high accuracy in the diagonal line indicates that the classifiers have converged on their own training set.}
\vspace{-2.2mm}
\label{tab:non_mutual_dg_imgnet}
\centering
\newcolumntype{x}{>{\centering\arraybackslash\hspace{0pt}}p{10.5mm}}
\footnotesize{
\begin{tabular}{l@{\hspace{1.5mm}}l@{\hspace{0mm}} x@{\hspace{-0.1mm}} x@{\hspace{-0.1mm}} x@{\hspace{-0.1mm}}  x }
\toprule
& \textbf{Source of Test Images} &Real &ADM & SG-XL \\
\midrule \midrule 
&\textbf{Classifier Trained on Real} &99.78 &71.11 &71.11 \\
&\textbf{Classifier Trained on ADM Images} &47.04 &99.69 &58.60 \\
&\textbf{Classifier Trained on SG-XL Images} &33.28 &46.77 &99.99 
\\
\bottomrule
\end{tabular}
}
\vspace{-1.3mm}
\end{table}
}

\newcommand{\tabdatasetinfoIMGNET}{
\begin{table}[!tbp]
\caption{The quantitative results of the sampled synthetic datasets from commonly used evaluation metrics: FID, Precision/Recall \cite{Kynkaanniemi2019}, and Diversity/Coverage \cite{ferjad2020icml}. Note that the reported FIDs were computed between the original dataset and the sampled synthetic dataset, therefore might differ from the reported FID of the checkpoints used for sampling.}
\vspace{-2.2mm}
\label{tab:datasetinfo_imgnet}
\centering
\newcolumntype{x}{>{\centering\arraybackslash\hspace{0pt}}p{12mm}}
\footnotesize{
\begin{tabular}{l@{\hspace{0mm}}l@{\hspace{0mm}} x@{\hspace{-1.9mm}} x@{\hspace{-0.1mm}} x@{\hspace{-0.1mm}} x@{\hspace{-0.1mm}} x@{\hspace{-0.1mm}} x@{\hspace{-0.1mm}} x }
\toprule
&\textbf{Dataset} &\textbf{Method}  &\textbf{FID}	&\textbf{Precision}	&\textbf{Recall}	&\textbf{Diversity}	&\textbf{Coverage}
   \\
\midrule \midrule 
&\multirow{2}{3.5em}{\textbf{ImageNet-10\%}}&{\cite{dhariwal2021diffusion}}  &4.91 &0.84 &0.61 &1.21 &0.94 \\
&&{\cite{Sauer2021ARXIV}} &5.28 &0.79 &0.58 &1.06 &0.90\\
\bottomrule
\end{tabular}
}
\vspace{-1.2mm}
\end{table}
}

\newcommand{\tabtrainvaltestRAND}{
\begin{table}[!tbp]
\caption{The achieved accuracy of classifiers (Random Initialization) trained and tested on different sources of images. Note that the validation and test sets containing only real CIFAR-10 images. The training sets are the one used to train the respective classifier.}
\vspace{-2.2mm}
\label{tab:train_val_test_rand}
\centering
\newcolumntype{x}{>{\centering\arraybackslash\hspace{0pt}}p{10.5mm}}
\footnotesize{
\begin{tabular}{l@{\hspace{1.5mm}}l@{\hspace{0mm}} x@{\hspace{-0.1mm}} x@{\hspace{-0.1mm}} x@{\hspace{-0.1mm}}  x }
\toprule
& \textbf{Source of Test Images} &Train &Val & Test \\
\midrule \midrule 
&\textbf{Classifier Trained on Real} &99.85 &91.13 &90.27 \\
&\textbf{Classifier Trained on LSGM Images} &99.83 &86.58 &85.08 \\
&\textbf{Classifier Trained on SG-XL Images} &99.96 &72.96 &72.22
\\
\bottomrule
\end{tabular}
}
\vspace{-1.3mm}
\end{table}
}

\newcommand{\tabtrainvaltestFT}{
\begin{table}[!tbp]
\caption{The achieved accuracy of classifiers (ImageNet Initialization) trained and tested on different sources of images. Note that the validation and test sets containing only real CIFAR-10 images. The training sets are the one used to train the respective classifier.}
\vspace{-2.2mm}
\label{tab:train_val_test_ft}
\centering
\newcolumntype{x}{>{\centering\arraybackslash\hspace{0pt}}p{10.5mm}}
\footnotesize{
\begin{tabular}{l@{\hspace{1.5mm}}l@{\hspace{0mm}} x@{\hspace{-0.1mm}} x@{\hspace{-0.1mm}} x@{\hspace{-0.1mm}}  x }
\toprule
& \textbf{Source of Test Images} &Train &Val & Test \\
\midrule \midrule 
&\textbf{Classifier Trained on Real} &96.34 &92.34 &91.80 \\
&\textbf{Classifier Trained on LSGM Images} &97.71 &86.10 &85.57 \\
&\textbf{Classifier Trained on SG-XL Images} &99.82 &79.43 &78.94
\\
\bottomrule
\end{tabular}
}
\vspace{-1.3mm}
\end{table}
}


%% file: main2.tex
\title{Improving the Effectiveness of Deep Generative Data}

\author{Ruyu Wang\textsuperscript{1,2} \quad Sabrina Schmedding\textsuperscript{1} \quad Marco F. Huber\textsuperscript{2,3}\\
\textsuperscript{1}Bosch Center for Artificial Intelligence, Renningen, Germany\\
\textsuperscript{2}Institute of Industrial Manufacturing and Management IFF, University of Stuttgart, Stuttgart, Germany \\
\textsuperscript{3}Fraunhofer Institute for Manufacturing Engineering and Automation (IPA), Stuttgart, Germany \\
{\tt\small \{ruyu.wang, sabrina.schmedding\}@de.bosch.com \quad marco.huber@ieee.org}
}
\maketitle

\begin{abstract}
Recent \acp{DGM} such as \acp{GAN} and \acp{DPM} have shown their impressive ability in generating high-fidelity photorealistic images. Although looking appealing to human eyes, training a model on purely synthetic images for downstream image processing tasks like image classification often results in an undesired performance drop compared to training on real data. Previous works have demonstrated that enhancing a real dataset with synthetic images from \acp{DGM} can be beneficial. However, the improvements were subjected to certain circumstances and yet were not comparable to adding the same number of real images.
In this work, we propose a new taxonomy to describe factors contributing to this commonly observed phenomenon and investigate it on the popular CIFAR-10 dataset. We hypothesize that the Content Gap accounts for a large portion of the performance drop when using synthetic images from \ac{DGM} and propose strategies to better utilize them in downstream tasks. 
Extensive experiments on multiple datasets showcase that our method 
outperforms baselines on downstream classification tasks both in case of training on synthetic only (Synthetic-to-Real) and training on a mix of real and synthetic data (Data Augmentation), particularly in the data-scarce scenario.
\end{abstract}


\section{Introduction}
\label{sec:introduction}

\figteaser

Over the past decade, \acp{DGM} powered by large-scale datasets have demonstrated revolutionary results in the field of image synthesis. Models based on different working theories such as the \ac{VAE} \cite{Kingma2013AutoEncodingVB} and \acp{GAN} \cite{goodfellow2020generative} have shown astonishing performance in generating realistic-looking images. More recently, diffusion-based models \cite{ho2020denoising, dhariwal2021diffusion} have emerged as the new state-of-the-art family in this challenging task and have broken the long-time dominance of the \ac{GAN} family. The rapidly improved image quality and mode coverage of generative models leaves us wondering: to which degree is synthetic data from these models ready to replace real samples for image recognition tasks, especially in the case of data scarcity?

Despite the inevitable performance degradation in low data regimes, many researchers have tried to adapt generative models to various domains where real data is hard and time-consuming to acquire (e.g., defective industrial products) \cite{Zhang_2021_WACV,9000806,Wang_2022_BMVC}. In such cases, images from generative models serve as a kind of data augmentation. For example, by navigating in the learned latent space, most of the generative models can deliver images with novel semantic information according to the guidance \cite{Collins2020EditingIS, jahanian2019steerability, 
Wu2020StyleSpaceAD, roich2021pivotal, Patashnik_2021_ICCV}. However, the usefulness of synthetic data is highly dependent on use cases as discovered in many works \cite{Jahanian2021GenerativeMA, sankaranarayanan2018learning, Zhang2021DatasetGANEL, Wang_2022_BMVC, Zhang_2021_WACV}. 
Tasks such as pose estimation \cite{Jahanian2021GenerativeMA} benefit more from synthetic data than object recognition tasks.
However, it has been widely observed that in most cases the effectiveness of synthetic images is not on par with that of real ones, and the boost brought by them not only saturates much faster but also diminishes rapidly once more real data are available \cite{Beckham2022ChallengesIL, Sariyildiz2022FakeIT, He2022IsSD}. Prior works generally attribute the cause of these effects to the \emph{Domain Gap}.

In this paper, we go beyond the general description of the \emph{Domain Gap} and define a new taxonomy to detail its possible factors into the \emph{Appearance Gap}, the \emph{Content Gap}, and the \emph{Quality Gap} as illustrated in Fig.~\ref{fig:teaser}. Then, we endeavor to answer the following questions:
\begin{itemize}
    \item Does the synthetic data generated by \acp{DGM} introduce novel information (rendering \acp{DGM} a semantically meaningful data augmentation method), or is its utility only equivalent to resampling a subset of the real dataset the \ac{DGM} was trained on?
    \item Which components of the \emph{Domain Gap} contribute most to cases with poor performance?
\end{itemize}
To address these questions, a novel set of investigations were conducted to examine the effectiveness of synthetic CIFAR-10 \cite{Krizhevsky09learningmultiple} images from two popular \acp{DGM} \cite{Sauer2021ARXIV, vahdat2021score}. Our results strongly suggest that the \emph{Content Gap} has the highest impact among the factors on the utility of synthetic images in downstream tasks. Inspired by these findings, we further propose two remedies---Pretrained Guidance and Real Guidance---, aiming to boost the effectiveness of synthetic data despite its current shortcomings in downstream recognition tasks. 

Extensive results on multiple datasets demonstrate that the proposed Pretrained Guidance, where we use a large-scale pretrained model (e.g., pretrained on ImageNet \cite{deng2009imagenet}) as a prior to regularize the distance between features during downstream classifier training, surpass other synthetic-to-real generalization methods. Moreover, together with Real Guidance to mitigate the negative effect of synthetic data, our method outperforms all baselines when mixing with real data, especially in the low data regimes.

\section{Related Work}
Domain adaptation \cite{ganin2015unsupervised, Tzeng2017AdversarialDD, dada} and generalization \cite{Li2017DeeperBA} have been important research topics in recent studies on deep learning, where the main aim is to recover performance drops caused by the domain gap. Among their sub-disciplines, synthetic-to-real adaption and generalization \cite{Richter_2016_ECCV, Peng2017VisDATV, Dosovitskiy2015FlowNetLO, Tobin2017DomainRF} have been promising research directions for domains where acquiring large amounts of data or labels is either hard or time-consuming (e.g.,  industrial, and medical). 
Early works mainly focused on synthetic images from \ac{PBR}, where the simulator aims to closely match the physical reality and achieve photo-realism. However, models trained on such kinds of synthetic images often generalize poorly on the target real data despite the relatively low cost of data collection. Many researchers have proposed different methods to mitigate the performance drop caused by the \emph{Appearance Gap} (e.g., unrealistic textures or over-simplified lighting conditions) such as by utilizing adversarial learning \cite{Muandet2013DomainGV, Yue2019DomainRA} or meta-learning \cite{Li2017LearningTG}. Lately, Chen et al. proposed to leverage the ImageNet pretrained weights as prior knowledge from real domain and combine it with a continual learning scheme \cite{chen2020automated} or a contrastive learning framework \cite{chen2021contrastive} to implicitly align learned features to their ImageNet counterparts. Unlike their works, we do not impose such feature similarity and instead relax the constraint by encouraging the similarity between the distance distribution of features.

More recently, \acp{DGM} have emerged as a new alternative to synthetic data generation \cite{Karras2018ASG, Brock2018LargeSG, Sauer2021ARXIV, vahdat2021score, ho2020denoising, dhariwal2021diffusion, xiao2021tackling}. They come with several favorable features over \ac{PBR}: 1) superior photorealism from training on real-world data; 2) requiring less storage space; 3) applicable for sophisticated cases that cannot be modeled by \ac{PBR} (e.g., defects in production).
Few earlier works\cite{Besnier2019ThisDD,  Jahanian2021GenerativeMA, Zhang2021DatasetGANEL, sankaranarayanan2018learning} have attempted to explore the synthetic data from \acp{DGM} for image recognition tasks but mainly focused on \ac{GAN}-based methods. 
With the recent bloom of diffusion-based models, some researchers have reported their investigation \cite{He2022IsSD, Sariyildiz2022FakeIT} regarding synthetic data from the state-of-the-art text-to-image generation models and revealed the potential of zero-shot generation, i.e., without seeing any real target data.  

However, it remains unclear how the synthetic data from \acp{DGM} trained on small-scale datasets (e.g., 100 or 1,000 samples) would affect the downstream recognition tasks.
Despite the appealing ability of zero-shot generation from popular text-to-image models, it is non-trivial to adapt it to some specific domains like industrial products, which are critically under-represented online. The plausible solutions therefore lie in low-/few-shot adaption \cite{Zhao2022ACL, wu2021stylealign, Robb2020FewShotAO, moon2022finetuning, NEURIPS2021_682e0e79, Giannone2022FewShotDM} of \acp{DGM}. Beckham et al. \cite{Beckham2022ChallengesIL} designed a shallow generative network to study the effectiveness of synthetic data in the few-shot scenario, where they concluded that the benefit of synthetic data was rather subtle in this case.
This phenomenon of synthetic data being less effective than real data has been widely observed in previous works \cite{Besnier2019ThisDD, Jahanian2021GenerativeMA, Zhang2021DatasetGANEL, sankaranarayanan2018learning} and only becomes more pronounced in the low data regime.
For scenery datasets, researchers \cite{Bau2018GANDV, Bau2019SeeingWA} have attributed the drop in performance to a distribution shift in the image content and reported that \acp{DGM} tend to omit rare or hard components (e.g., humans). We would consider this one aspect of the \emph{Content Gap} and investigate it for object-centered images. Unlike scenery images, the loss of a mode representing an object with plausible but rare attributes (e.g., a cat in the water) might be easily overlooked. In this work, we study the impact of such mode drop on downstream classification tasks and propose remedies based on our findings.

\section{Empirical Investigations}
\label{sec:observations}
In this section, we first introduce a new taxonomy to detail the potential factors for the \emph{Domain Gap} and then present an empirical investigation of the two questions stated in Sec.~\ref{sec:introduction}, linking the 
 factors of \emph{Domain Gap} to the effectiveness of synthetic data. 

\subsection{Domain Gap Factors}
Researchers often refer to the negative effects, which happen when the data distribution of training and testing sets are different, as \emph{Domain Gap}. 
However, it remains unclear what exactly are the factors leading to the distribution shift. To dig deeper into multiple constituents of the domain gap, we classify the potential factors into three categories---the \emph{Appearance Gap}, the \emph{Content Gap}, and the \emph{Quality Gap}. The \emph{Appearance Gap} refers to the performance gap induced by artifacts such as unrealistic textures, which is more often observed in the synthetic data acquired from \ac{PBR} \cite{pharr2016physically}. 
The \emph{Content Gap} addresses the distribution shift in the composition of a generated image. For example, Bau et al. \cite{Bau2019SeeingWA} studied the typical failure cases of \acp{GAN} by examining the deviation between a specific scene image and its reconstruction, concluding that the models tend to skip difficult subtasks like large human figures. Finally, the \emph{Quality Gap} is stated for the structural distortion in the synthetic images such as two-headed fish or faceless humans.

\subsection{Study on Synthetic Datasets}
Our exploration on effects of synthetic data was carried out by first assuming full access to a widely used benchmark dataset in image synthesis---CIFAR-10, and checkpoints of popular \acp{DGM} trained on CIFAR-10. Then, we sampled synthetic clones of CIFAR-10 from the selected \acp{DGM} and analyzed the behavior of classifiers trained on the original and synthetic sets of CIFAR-10 for object-centered image classification. 
Analogous experiments and effects on ImageNet can be found in Appendix~\ref{appendix:study_imgnet}.



\textbf{Experiment setup.} 
We selected two popular \acp{DGM} with state-of-the-art FID \cite{Heusel2017GANsTB} at the time---LSGM \cite{vahdat2021score}
and StyleGAN-XL (SG-XL) \cite{Sauer2021ARXIV}---and sampled images directly from the provided checkpoints by the authors.  
LSGM is limited to unconditional image synthesis. We therefore leveraged an off-the-shelf Wide Residual Network (WRN) \cite{Zagoruyko2016WideRN} (Top-1 Accuracy on CIFAR-10 classification: 96.21\%\footnote{https://github.com/xinntao/pytorch-classification-1}) during the sampling process to obtain labels for the synthetic dataset (with predicted softmax probability $>$ 0.8). 
To make a fair comparison, we also applied the same filtering procedure on the conditional SG-XL.

\tabnonmutualdg
\figtraincurves
We present the results under the setting that all classifiers were initialized with pretrained ImageNet weights and the learning rate was set to 0.0001 (See Appendix~\ref{appendix:rand_eir} for random initialization). 
Although the performance drops are extensively observed in different network structures (e.g., EfficientNet \cite{tan2019efficientnet}, ViT \cite{dosovitskiy2020image}), we chose the widely deployed ResNet-50 \cite{He2015DeepRL} as backbone for demonstration. 
We set the image resolution to 224 $\times$ 224, the batch size to 128 and trained the models for 200 epochs. To observe the full effect of synthetic data, only random crop and random flip were applied as data augmentation. 
We evaluated all models on the official CIFAR-10 test set and reported the averaged accuracy over five random runs. 
Note that we employed a fixed train-val split for the real dataset throughout the paper---500 samples from each class were randomly selected in the beginning as the validation set and the remaining 45,000 images formed the training set. 
For simplicity, we later denote real images as \emph{Real} and synthetic images from \acp{DGM} as \emph{Synthetic}.

\textbf{Observation \#1: The \emph{non-mutual} performance gap.} 
It has been widely observed that there exists a performance gap between real and synthetic data. A commonly accepted hypothesis attributes this behavior to the domain gap between the two distributions. 
Based on this assumption, one would presume that the domain gap between synthetic data and real data is mutual, meaning that the performance drop in the case of synthetic-to-real should be at a similar scale to the opposite case of real-to-synthetic. However, this presumption does not hold as seen in Tab.~\ref{tab:non_mutual_dg}. We can see that the performance of the classifier trained on \emph{Real} is not affected too much when classifying synthetic images, while the classifiers trained on \emph{Synthetic} sources (e.g., SG-XL) exhibit drastic degradation in accuracy. Moreover, SG-XL shows a more severe decrease than LSGM, presumably due to diffusion models being better in mode coverage than \acp{GAN}.

\figfakelossdistr

\textbf{Observation \#2: The training accuracy saturates quickly when training on \emph{Synthetic} sources.} 
We show the training and validation accuracy curves of all image sources in Fig.~\ref{fig:training-curves}. It can be seen that the training accuracy of \emph{Synthetic} sources grew much faster in the early epochs ($<$ 25) and saturated faster compared to the classifier trained on the \emph{Real}. While the performance gap between the training and validation can be explained by the distribution shift, the peculiar behavior of the training curves of \emph{Synthetic} was puzzling---all the classifiers were tasked with the same objective (i.e., classify the images into ten classes), why would the classifiers trained on the  \emph{Synthetic} sources seemed to be solving an easier task? 
We hypothesize that this behavior can be attributed to the \emph{Content Gap} between synthetic and real data, where the compositions of an object and its rare but possible attributes (e.g., a car in the sea) were either dropped by the models or barely generated. Therefore, the resulting synthetic dataset only contains images associated with frequent attributes, forming a less diverse and simplified training set.
This led to the next question: do the current synthetic images from the \acp{DGM} just reflect a subset of the real data?

\textbf{Observation \#3: Very small losses are observed for \emph{Synthetic} samples.}
To validate the hypothesis that synthetic images only cover a subspace of the real dataset, we further examine the training samples from all sources. We propose to measure the importance of a sample by the cross entropy loss it brings to a trained classifier of an opposite source. This means we evaluate all the samples from the \emph{Synthetic} sources by a classifier trained on \emph{Real} samples and vice versa. The concept behind it is that if an incoming sample contains novel information which was not captured during the training process, the loss it simulates should be high. 
Conversely, if the information a sample can bring is already learned by the model, its loss should be close to zero.

We plot the measured losses in Fig.~\ref{fig:fake-loss-distr}. In each subgraph, the solid bars present the loss distribution in the case where the classifier was trained on \emph{Real} data and used to evaluate \emph{Synthetic} images. On the contrary, the dotted bars show the case of training a classifier on \emph{Synthetic} and test on \emph{Real} images. 
It can be observed that the synthetic samples (Solid) overall have a lower loss compared to the real ones (Dotted). Compared to the losses brought by the real samples (Dotted), much more synthetic samples (Solid) have a close-to-zero loss and only a small portion possess a value bigger than 1 (See more low- and high-loss images in Appendix~\ref{appendix:loss_images}).
Interestingly, we still observe that utilizing the low-loss synthetic samples for data augmentation is beneficial for downstream performance (see Appendix~\ref{appendix:ft_eir}). 

\subsection{Conclusion}
All three observations emphasize that despite training on a relatively large dataset and having promising FID, images from \acp{DGM} still suffer from the domain gap. 
While the \emph{Appearance Gap} is eliminated by learning directly on the real data, it was unclear whether the synthetic images in our investigations suffered from \emph{Quality} or \emph{Content Gap}.
The following two details however suggest that the \emph{Content Gap} is the main issue:
(1) The accuracy of \emph{Real} classifiers in \textbf{Observation \#1} does not show signs of obstruction, as would be assumed from distorted images.
(2) The distorted data should also simulate a high loss in \textbf{Observation \#3}, yet the loss from synthetic samples is generally lower than that of the real ones.

We thus conclude that sampling directly from current \acp{DGM} will result in a subset of the real data due to the \emph{Content Gap}, which worsens the long-tailed nature that most real-world data exhibit---rare samples are either omitted by the generative models or barely ever generated. Therefore, downstream feature extractors trained exclusively on such misrepresented distributions tend to show undesirable behavior (e.g., performance drop in classification accuracy) when tested on real data.
However, at the end of \textbf{Observation \#3}, we observe that even adding synthetic samples with lower loss to the real training set still has minor positive effects. We hypothesize that this is because the synthetic images from \acp{DGM} are lacking rare samples but do add variations within the covered parts of the training distribution.
As a result, we introduce two strategies to boost their effectiveness in the following section. 

\section{Remedies}
\label{sec:remedies}
The \emph{Content Gap} can also be cast as a mode coverage problem: sampling synthetic data most likely leads to a less diverse and simplified dataset, 
where rare samples are not represented. The information such a simplified dataset can provide is therefore mostly covered by the original dataset. 
Unsurprisingly, training recognition models on such a dataset can easily result in degraded performance and factitious-biased representation.
One potential way to acquire synthetic data with higher information density is to apply additional techniques during the sampling process \cite{Humayun2022PolaritySQ, Han2022RarityS, Watson2022LearningFS}. However, this only works under the assumption that the rare samples are still learned by the \acp{DGM} and are just rarely generated. The assumption might be violated when the \acp{DGM} are trained on small datasets and the rare cases in the training set are omitted by the models. 
This is a practically relevant task however, because it is often impossible to acquire large-scale data for training a well-behaved \ac{DGM} from scratch in most real-world scenarios (e.g., industrial). As a result, we propose two novel regularization strategies  to improve the effectiveness of the synthetic data even in low data regimes, termed as \emph{Pretrained Guidance} and \emph{Real Guidance}. 


\figpgrg

\textbf{\ac{PG}}. Models pretrained on large-scale datasets like ImageNet often exhibit diverse and rich representations that can be transferred to other tasks. 
We propose exploiting this trait as external guidance to prevent the models trained on the synthetic dataset from converging into having less distinctive representations.
The closest previous works \cite{chen2020automated, chen2021contrastive} suggested tackling this problem by forcing the new model to have a similar probability distribution to the pretrained ImageNet predictions, implicitly regularizing the new feature extractor to stay close to the ImageNet. 
But such constraints are not always beneficial and their performance depends heavily on how the model is initialized as we later show in Sec.~\ref{sec:experiments}.
In contrast, our proposed \ac{PG} allows the feature extractor to learn freely from the new dataset while maintaining a similar span of data as ImageNet.

The workflow of our \ac{PG} is illustrated in Fig.~\ref{fig:pgrg}(a). We denote the model being trained as $M_{u}$ and utilize a frozen pretrained model $M_{p}$ (note that $M_{p}$ can be different to $M_{u}$). 
In each forward pass, a synthetic sample will be given to both networks and obtain the feature representations $f_{p}$ and $f_{u}$, respectively. For a batch of $N$ images, the distance matrices $D_{p} \in \mathbb{R}^{N \times N}$ and $D_{u} \in \mathbb{R}^{N \times N}$ are computed with respect to $f_{p}$ and $f_{u}$ using a metric $\delta$. We then propose to encourage $D_{p}$ to possess a similar distance distribution as $D_{u}$.
In this way, $f_{u}$ is allowed to learn the representations for the target task freely, while the guidance from the pretrained model helps preserve the span of features.
We formulate the proposed \ac{PG} regularization as 
\begin{equation}
\begin{split}
& D_{p}^{ij} = \delta \big(f_{p}(i), f_{p}(j) \big), \; D_{u}^{ij} = \delta \big(f_{u}(i), f_{u}(j)\big)~, \\
&\mathcal{L}_{\text{pg}} = \sum\nolimits_{i}  \sum\nolimits_{j} \text{sim} \bigg\{ D_{p}^{ij}, \; D_{u}^{ij}\bigg\}~, 
\end{split}
\end{equation}
where $i$, $j$ indicate different samples in a batch, 
$\delta(\cdot)$ stands for any distance metric like the cosine similarity, and sim\{$\cdot$\} can be any similarity metric such as L1 or KL-divergence\footnote{KL-divergence on distance  matrices as derived in \cite{ojha2021few}}.
Note that the effect of \ac{PG} is independent of initializing the model with pretrained weights because despite providing a good starting point, the initialization does not stop the model from converging into having less distinctive representations.

\textbf{\ac{RG}}.
Inspired by replay-based continual learning (CL) approaches \cite{Chaudhry2018EfficientLL}, we adapt the idea of gradient episodic memory to our context to mitigate the domain gap. In contrast to alleviating forgetting, we utilize it as a way to regularize the gradient flow of the synthetic data. Specifically, we assume that a small set of real data (e.g., 10 or 100 samples) is available at training time. It is a more realistic scenario than assuming the sole availability of synthetic data, regarding it would usually require a handful of real data to train or finetune the DGMs in the beginning.

Considering the gradients $g_r$ from real data as a positive influence on the model, the gradients $g_f$ from synthetic data can however be a confounder if the angle between $g_r$ and $g_f$ are more than $90^{\circ}$.
We then channel the gradients from the synthetic data so that they are at most perpendicular to the real data, avoiding the misleading signal brought by $g_f$. 
This can be formulated as 
\begin{equation}
\begin{split}
&g_{f_\text{new}}= 
\begin{cases}
      g_f - \frac{g_f^{\top}g_r }{g_r^{\top}g_r } \cdot g_r&, \text{if } g_f^{\top}g_r < 0\\
      g_f &, \text{otherwise}
\end{cases}
, 
\\ &g_{\text{update}}= \lambda_{1} \cdot g_{r} + \lambda_{2} \cdot g_{f_\text{new}}~.
\end{split}
\end{equation}
In this way, we prevent the synthetic data from overfitting itself while narrowing the domain gap by introducing information from the real data.

In summation, the final objective for a target task is
\begin{equation}
\begin{split}
&\mathcal{L} = \mathcal{L}_{\text{Task}} + \lambda_{3} \cdot \mathcal{L}_{\text{pg}}~,
\end{split}
\end{equation}
where $\mathcal{L}_{\text{Task}}$ can be a cross-entropy loss in case of classification and $\lambda_{1}$, $\lambda_{2}$, $\lambda_{3}$ are the hyperparameters to tune the intensity of our proposed regularizations.


\section{Experiments}
\label{sec:experiments}
To study the effectiveness of synthetic data from popular \acp{DGM} and to evaluate the usefulness of our proposed remedies against the expected content gap, we consider two cases: synthetic-to-real generation and synthetic data as data augmentation. The first case represents a scenario in which no real data is available for downstream tasks, which we also refer to as the \textbf{Zero-shot} task. Although it is a theoretical setting ignoring the real samples that were required to train a \ac{DGM}, the results from this case can serve as a quantitative assessment of the discrepancy between synthetic and real data. 
The second case presents a scenario assuming at least a handful of real images (e.g., 100 or 1,000) are available, which we describe as the \textbf{Low-shot} task. It is more common in real-life settings (e.g., industrial), otherwise training or finetuning a \ac{DGM} as well as evaluating in downstream tasks would be generally infeasible. 

\textbf{Datasets.} Besides the three sets of CIFAR-10 (\emph{Real}, \emph{LSGM} and \emph{SG-XL}) mentioned in  Sec.~\ref{sec:observations}, synthetic images of two popular benchmark datasets (CUB Bird \cite{WahCUB_200_2011}, Oxford Flower \cite{Nilsback08}) and one industrial dataset (SDI \cite{Wang_2022_BMVC}\footnote{We thank the authors \cite{Wang_2022_BMVC} for providing the dataset for our benchmark.}) from three generative models\footnote{Diffusion-based models are omitted due to their low performance in the low data regimes. See Appendix~\ref{appendix:dgm_details} for more information.} (SG-XL, Projected GAN (Proj-GAN) \cite{Sauer2021NEURIPS}, and DT-GAN \cite{Wang_2022_BMVC}) were used for evaluations. For the first two datasets, we sampled images to meet the size of the original datasets. For the SDI dataset, we sampled images to balance the original SDI dataset and thus, resulted in three sets of synthetic data with size 8,000.
Note that due to the mode-dropping issue when training on small datasets, we conducted all classification experiments on a subset of CUB Bird and Oxford Flower. 
Details on generative model training and sampling are in Appendix~\ref{appendix:dgm_details}. 


\tabotherfakes

\textbf{Experiment Setup.}
We deployed the same setup for training ResNet-50 classifiers as described in Sec.~\ref{sec:observations}.
All following experiments were conducted under the same setting unless otherwise specified. The reported accuracies were averaged over five random runs. We used default hyperparameters from the authors for selected baselines and set $\lambda_{1}$, $\lambda_{2}$ and $\lambda_{3}$ of our method differently for each dataset. See Appendix~\ref{appendix:hyper_params} for the chosen hyperparameters and Appendix~\ref{appendix:more_results} for more results in different settings.

\subsection{Zero-shot Image Classification}


We investigate the zero-shot classification performance of the synthetic data from various models. As seen in Tab.~\ref{tab:other_fake_cls}, simply training the classifiers on synthetic images leads to a degraded performance compared to the classifiers trained on the same amount of real data. 
However, has the current synthetic data reached its full potential? We argue that the drop in performance can be reduced based on our insights from Sec.~\ref{sec:observations} and the proposed \ac{PG} for regularization as introduced in Sec.~\ref{sec:remedies}. 
It can be seen that methods like \cite{chen2020automated, chen2021contrastive} highly depend on the initialization of the model. Also, their regularizations are not always beneficial. For example, they often leads to worse results when the number of samples in each class is relatively high (e.g., CIFAR-10). For datasets that do not benefit from the ImageNet initialization even for real data (e.g., industrial products like SDI-A and SDI-C in Tab.~\ref{tab:other_fake_cls}), these methods often fail to deliver promising results. 
In contrast, our \ac{PG} provides better performance in most cases considering both kinds of initialization, suggesting its practical usefulness in real-world industrial cases. 


\tabcifarsmall

Moreover, our \ac{PG} requires not the same model architecture for the frozen pretrained model $M_p$ and the new model $M_u$. Therefore, it can also combine pretrained weights from wide-spread architectures to new models where no pretrained weights are available.
As an exemplary case, we set $M_u$ to a simplified ResNet-50\footnote{https://github.com/kuangliu/pytorch-cifar}, where its network structure was customized to work for CIFAR-10 at resolution 32 $\times$ 32 and its ImageNet pretrained weights were not available.
Meanwhile, $M_p$ was set to be the original ResNet-50 with pretrained weights from ImageNet. It should be noted that the feature representations $f_p$ and $f_u$ have different dimensions (1024 vs. 256) due to the structural difference.
As shown in Tab.~\ref{tab:cifar_small}, our method again outperforms ASG in this case and delivers the best results when using the KL distance. CSG on the other hand is not feasible in such a scenario due to its design requiring $M_p$ and $M_u$ to have identical structures. 

The flexibility of our method also allows choosing the distance and similarity metrics freely. Our results have shown that applying KL distance together with cosine similarity generally yields better performance in our experiments. All further experiments were therefore conducted with this setting only.

\tabrealshots

\subsection{Low-shot Image Classification}
In the low-shot scenario, we investigate our method in the situation where the synthetic data is used as data augmentation to enlarge the real training set.
We also compare our results to baseline methods that were designed for domain adaptation, where the synthetic data and real data are used sequentially instead of simultaneously.  

We conducted the experiments in the following two settings: 
(1) Assuming that a large-scale dataset (CIFAR-10) was available for  pretraining generative models.
For the downstream task, classification in our setting, only very few real samples ($<$ 20) from the large-scale dataset are available.
(2) Assuming that a smaller scale dataset (e.g., CUB, Flower, and SDI) is available but has no public accessible pretrained generative models. The training data therefore needs to be used for both to train the generative models from scratch and to train the model for the downstream (classification) task. 
Note that these settings result in datasets with different proportions of synthetic and real images, which we denote as ``Syn-to-Real Ratio''.

Theoretically, more  settings would be possible such as using pretrained weights from a large-scale dataset for finetuning the generative model on a smaller scale dataset. However, finetuning a \ac{DGM} requires carefully adaptions and 
we leave a complete study of all options for future work.

\textbf{Baselines.} 
For data augmentation, we set the baseline as simply adding the synthetic images to the real training set. In addition, we selected A-GEM\cite{Chaudhry2018EfficientLL} as regularization to aid this baseline. Conceptually, this is the closest baseline to our method. 
For domain adaptation, we chose four domain adaptation methods as baselines---DANN \cite{ganin2015unsupervised},ADDA \cite{Tzeng2017AdversarialDD}, DADA \cite{dada} and LTDA \cite{JamalLongtail_DA}. Note that unsupervised methods like ADDA and DADA do not use the labels of real images. Also, all methods except DANN deployed a two-stage training scheme, meaning the model is first pretrained on synthetic data for 150 epochs and then finetuned on real data for 50 epochs instead of directly training for 200 epochs.
To incorporate real data during the training process, DANN, ADDA, and DADA aim to learn a domain-agnostic feature extractor via an adversarial scheme. LTDA deploys a meta-learning framework to adapt the model to the target distribution with a small subset from the target domain.

We report the results for different datasets in Tab.~\ref{tab:real_shots}. 
It is worth noting that the domain adaptation methods besides DANN drastically underperformed compared to the baseline data augmentation. This is predictable because all the domain adaptation methods except DANN deployed a two-stage training scheme, where the synthetic data used for pretraining are not directly accessible at the finetuning stage. However, it also shows that the current domain adaptation methods, which mainly tackle the significant distribution shift (e.g., cartoon dogs $\leftrightarrow$ dogs in natural images), do not address the fine-grained distribution shift between the synthetic images from a \ac{DGM} and the real images. We believe that such kind of synthetic-to-real adaptation could be an interesting topic for future work. 

However, using synthetic data for data augmentation is a promising approach. 
Note that the one-stage training scheme allows DANN to access synthetic and real data during the training process directly. 
Therefore, its performance is closer to the data augmentation baseline.
We also compared our results to A-GEM, where the gradients from the real images are only used to constrain the gradients from the synthetic ones but not used to update the model. 
We argue that in the case of synthetic data as data augmentation, it is important to also incorporate the real gradients during the training, particularly when the available real data are more than a few shots. 
It can be observed in Tab.~\ref{tab:real_shots} that applying our proposed \ac{PG} and \ac{RG} largely improved the classifier performance in most cases, especially in the low data regimes.
This showcases the effectiveness of our method in regularizing the distances between data points and eliminating the negative effect of the synthetic data.

\section{Conclusion}
In this work, we examined to which degree synthetic data from popular \acp{DGM} can replace real data in downstream recognition tasks. While a severe performance drop is commonly observed in the context and generally attributed to \emph{Domain Gap}, we further classified the potential factors into three categories. Moreover, we presented a series of observations indicating that the performance degradation can be mainly attributed to the \emph{Content Gap}, where the synthetic data from \acp{DGM} can only form a simplified dataset in which rare samples are not represented. 
Motivated by this hypothesis, we proposed a novel method combining two strategies---Pretrained Guidance and Real Guidance---to regularize the downstream models to keep the span of features despite training on a simplified dataset, even when the regularization comes from a different network architecture.
Extensive results on multiple datasets show that our method not only outperformed other synthetic-to-real generalization methods in zero-shot scenarios but also largely improved the effectiveness of the synthetic data when serving as data augmentation in low-shot settings.

%% file: bibcommand.tex
{\small
\bibliographystyle{ieee_fullname}
\bibliography{egpaper}
}

%% file: appendix.tex
\appendix

\section{Implementation Details}
In this section, we will provide the full details for our experiments in Sec.~\ref{sec:experiments} for reproducibility.

\subsection{Training and Sampling from Deep Generative Models for Different Datasets}
\label{appendix:dgm_details}
\subsubsection{CIFAR-10}
We used the checkpoints provided in the official repositories, where LSGM\footnote{https://github.com/NVlabs/LSGM} was reported with FID 1.94 and StyleGAN-XL\footnote{https://github.com/autonomousvision/stylegan-xl} with FID 1.85. LSGM only supports unconditional training on CIFAR-10. Therefore, we additionally applied a classifier trained on CIFAR-10 for labeling and rejection sampling. A Wide Residual Network (WRN) was chosen for this classification task, where the structure \emph{WRN-28-10} was selected. We used the checkpoint provided from an open repository\footnote{https://github.com/xinntao/pytorch-classification-1}, where an accuracy of 96.21\% on CIFAR-10 was reported. For rejection sampling, we set the threshold to $0.8$, meaning that we exclude all the samples with a prediction probability lower than 80\%. The same classifier was applied to StyleGAN-XL as a filtering mechanism on top of the class-conditional sampling to allow for a fair comparison.

\subsubsection{CUB-Bird and Oxford-Flower}
Compared to CIFAR-10, CUB-Bird and Oxford-Flower are much smaller datasets according to their total size (4,521 and 1,010) and the average number of samples per class (22 and 10). This poses a challenging task for training deep generative models, especially for the diffusion-based kind. 
We trained four models (details below) on both datasets from scratch, following the instructions from the respective authors' official repositories. All models were trained to generate images at resolution 256 $\times$ 256.

\textbf{LSGM.} We followed the instructions for ``CelebA-HQ-256 Quantitative Model" in the original repository and modified the command to train on one Tesla V100 GPU.

\textbf{Fast-GAN.} Following the instructions in the original repository\footnote{https://github.com/odegeasslbc/FastGAN-pytorch}, we trained the models on both datasets on one Tesla V100 GPU with batch size 12 for 100,000 iterations.

\textbf{Projection-GAN.} We trained the models on both datasets with one Tesla V100 GPU with batch size 8 for 20,000 kimgs (i.e., the model went through this number of images\cite{karras2019style}). Note that for CUB-Bird, we used the configuration of \emph{fastgan} while for Oxford-Flower we applied \emph{fastgan-lite} according to author's suggestions based on dataset size. 

\textbf{StyleGAN-XL.} Following the instructions in the original repository, we first trained a model for each dataset at resolution 32 and then directly scaled it up to resolution 256 at the second stage of training. All the models were trained on one Tesla V100 GPU with batch size 8 for 10,000 kimgs.

For sampling, we deployed a transformer-based classifier structure---Big Transfer (BiT) for labeling and rejection sampling. We picked the architecture of \emph{BiT-M-R50x1} as the backbone. The BiT classifiers for both datasets were trained from scratch and reached the accuracy of 82.52\% and 98.26\% on the validation set for CUB-Bird and Oxford-Flower, respectively. The threshold for rejection sampling for both datasets was set to $0.5$.

To construct a synthetic version of both datasets, we aimed to sample 30 images for each class in CUB-Bird and 10 images for each class in the case of Oxford-Flower and let the rejection sampling process run for at most 10 days.
We report the training and sampling statistics in Tab.~\ref{tab:dgm_training}.
Among all the models, only \emph{LSGM} CUB failed to converge. \emph{Proj-GAN} Flower was the only model to reach the target amount of images we seek to have within the 10-day timeframe. As a result, we only selected the synthetic datasets generated from \emph{Proj-GAN} and \emph{SG-XL} for further experiments in the main paper. Also, we omitted six classes (i.e., \emph{Least Auklet}, \emph{Spotted Carbird}, \emph{Northern Flicker}, \emph{Slaty Backed Gull}, \emph{Whip Poor Will} and \emph{Bohemian Waxwing}) in CUB-Bird and one class (i.e., Tiger Lily) in Oxford-Flower, forming a subset of the original dataset---194 classes and 101 classes for CUB-Bird and Oxford-Flower, respectively.

\tabdgmtraining

\subsubsection{SDI}
As for the SDI dataset, we thank the authors of DT-GAN\cite{Wang_2022_BMVC} for providing us the real dataset and the synthetic dataset, where the statistics of the real dataset can be found in the supplementary of their paper and the synthetic dataset contains an 8,000-image subset for each product.

\subsubsection{Statistic of the Sampled Synthetic Datasets}
We report the statistics of the sampled sythetic datasets in Tab.~\ref{tab:datasetinfo}. Interestingly, we note that FID can serve as a rough indicator for the effectiveness of the synthetic datasets, but how to predict the precise effect of these images in downstream tasks is yet to discover. Additionally, the scores of recall, which measures the fraction of the training data manifold being covered by the generator, are the most indicative among the other three metrics. We interpret it as a support sign for our claim---that the \emph{Content Gap} (i.e., mode coverage) exhibited in the synthetic datasets is a main factor for the performance drop in downstream tasks.

\tabdatasetinfo

\subsection{Hyperparameters for Pretrained Guidance and Real Guidance}
\label{appendix:hyper_params}
\tabotherfakesHP
We report the selected hyperparameters for \textbf{Zero-shot} classification in Tab.~\ref{tab:other_fake_cls_hp}.  For Tab.~\ref{tab:cifar_small} in the main paper, we used $\lambda_{3}=10$ for L1 distance and $\lambda_{3}=1,000$ in the case of KL-divergence. We observed that stronger regularization is in general needed when using random initialization.
For \textbf{Low-shot} classification, we report the used hyperparameters in Tab.~\ref{tab:real_shots_hp}. Note that when the number of available real images is much lower than the number of synthetic images, we empirically found that not updating the model with the gradients from real data (e.g., $\lambda_{1}=0$ for CIFAR-10) led to better performance.

\tabrealshotsHP

\section{Extended Investigation Results}
In this section, we present additional results of the empirical investigations in Sec.~\ref{sec:observations}.

\tabtrainvaltestFT
\subsection{Results from ImageNet Initialized Classifiers}
\label{appendix:ft_eir}
We present the achieved accuracy of the
classifiers on their respective training sets as well as on the
real CIFAR-10 validation and test set in Tab.~\ref{tab:train_val_test_ft}. 
Additionally, we sorted the \emph{Synthetic} images based on their sample losses in \textbf{Observation \#3} and divided them into two subsets of equal amounts. Then, we combined each subset with the \emph{Real} images to form two augmented new training sets and trained new classifiers on top. The results shown in Tab.~\ref{tab:fake_subsets} endorse the legality of judging samples by their loss, as it can be seen that the subsets with larger losses boost the performance more than the smaller halves. Note that despite having small losses, the subsets of such images still have a positive impact on the performance. 
We hypothesize that this is because the synthetic images from \acp{DGM} do add variations in the dense areas of the training distribution despite lacking rare samples.

\tabfakesubsets

\tabtrainvaltestRAND
\subsection{Results from Randomly Initialized Classifiers}
\label{appendix:rand_eir}
For the setting where all classifiers were randomly initialized, the learning rate was set to 0.01 while all other hyperparameters remained the same as in the main paper. We report the achieved accuracy of the
classifiers on their respected training sets as well as on the
real CIFAR-10 validation and test set in Tab.~\ref{tab:train_val_test_rand}.
\tabnonmutualdgRand
It can be seen in Tab.~\ref{tab:non_mutual_dg_rand} that the non-mutual performance gap (\textbf{Observation \#1} in the main paper) is even more pronounced with random initialization.
\figtraincurvesRand
Also, the saturation effect on training and validation curves as mentioned in \textbf{Observation \#2} can be clearly observed in Fig.~\ref{fig:training-curves-rand}.
\figfakelossdistrRand
Last but not least, the resulting loss distributions plots  in this setting (cf. Fig.~\ref{fig:fake-loss-distr-rand}) show the same tend as discovered in \textbf{Observation \#3} and the classifier performance in Tab.~\ref{tab:fake_subsets_rand} consists with our finding in Tab.~\ref{tab:fake_subsets}. 
We therefore conclude that our insights in Sec.~\ref{sec:observations} are non-negligible and independent from the initialization method.
\tabfakesubsetsRand

\subsection{Visualization of High-loss and Low-loss Images}
\label{appendix:loss_images}
We show the high-loss and low-loss \emph{Synthetic} images in Fig.~\ref{fig:fake-lsgm} and Fig.~\ref{fig:fake-sgxl}, and the \emph{Real} images evaluated by classifiers trained on  \emph{Synthetic} images in Fig.~\ref{fig:real-lsgm} and Fig.~\ref{fig:real-sgxl}. It can be observed that the quality of the \emph{Synthetic} images from \ac{DGM}s is on par with the \emph{Real} images from the original CIFAR-10 training set. However, compared to the high-loss \emph{Synthetic} samples, the high-loss \emph{Real} samples (i.e., the hard cases in view of classifiers trained on \emph{Synthetic} data) resemble rarer but plausible attributes (e.g., the dog with red hat in Fig.~\ref{fig:real-sgxl}). We interpret this as a support sign of our claim in the main paper---that the \emph{Synthetic} dataset is less diverse and simpler than its original training set due to the absence of rare samples.

\subsection{Study on ImageNet}
\label{appendix:study_imgnet}
To demonstrate that the effects we observed in Sec.~\ref{sec:observations} are not confined to CIFAR-10, we also conducted the same investigation on a subset of ImageNet \cite{deng2009imagenet}---ImageNet-10\%, where 128 samples of each class were randomly selected to form the subset. This resulted in a training set of 128,000 samples and we reported the statistics of the sampled synthetic  datasets in Tab.~\ref{tab:datasetinfo_imgnet}. For evaluation, we split the official ImageNet validation set into a validation set of size 12,000 and a test set of size 38,000.

\tabdatasetinfoIMGNET

\textbf{Experiment setup.} We selected two popular \acp{DGM}---ADM \cite{dhariwal2021diffusion} and SG-XL \cite{Sauer2021ARXIV}---based on their promising performance on ImageNet and sampled directly from the provided checkpoints by the authors. We used the ImageNet class index as the condition to sample 128 images for each class at resolution 256 $\times$ 256 from both \ac{DGM}.

Same as in the main paper, we chose ResNet-50 \cite{He2015DeepRL} as the backbone for the classifiers and set the image resolution to 224 $\times$ 224, following the common preprocessing procedure for ImageNet.
The batch size was set to 256 and four NIVIDA Tesla V100 were used. The initial learning rate was set to 0.1 and a cosine annealing schedule was applied to tune the learning rate during the training. Only random crop and random flip were used as data augmentation. Note that we randomly initialized the network weights to observe the full effect of synthetic data. Initializing with ImageNet pretrained weights as in the main paper would eliminate the need for classifier training since the target dataset is also ImageNet.
All the classifiers were trained for 300 epochs and the reported results in the following were acquired from averaging over three random runs.

\tabtrainvaltestIMGNET

\textbf{Observations.} We present the achieved accuracy of the classifiers on their respective training sets as well as on the real ImageNet validation and test set in Tab.~\ref{tab:train_val_test_imgnet}. Together with Fig.~\ref{fig:training-curves-imgnet}, a sign of underfitting can be observed even on the classifier trained with real data, presumably due to the reduced size of the training set. However, we believe the experiments can still serve as a proxy for the behavior of the full dataset. As shown in Tab.~\ref{tab:non_mutual_dg_imgnet}, the non-mutual performance gap we claimed in Sec.~\ref{sec:observations} (\textbf{Observation \#1}) can be clearly observed. Note that despite the notable drop when applying the classifier trained on \emph{Real} to the \emph{Synthetic} samples, the achieved accuracy is still significantly higher than on the test set of ImageNet (71.11\% $\rightarrow$ 57.67\%). Also, it can be seen in Fig.~\ref{fig:training-curves-imgnet} that the training curves show the same trend as in Fig.~\ref{fig:training-curves}, fitting our claim in \textbf{Observation \#2} that the training accuracy saturates quickly when training on \emph{Synthetic} sources. Finally, we plotted the loss distribution in Fig.~\ref{fig:fake-loss-distr-rand-imgnet} and visualized the low-loss and high-loss samples in Fig.~\ref{fig:fake-sg-imgnet}-\ref{fig:real-gd-imgnet}, where the same conclusion as in \textbf{Observation \#3} can be drawn---that the datasets formed by \emph{Synthetic} samples contain less information compared to the real one.

\tabnonmutualdgIMGNET
\figtraincurvesIMGNET

We interpret all of these observed effects as support signs to our claim that the synthetic datasets from current \acp{DGM} are the simplified version of the original dataset, where we assume the reason to be that the rare samples are either lost or under-represented in the sampled sets (i.e., the \emph{Content Gap}). 

\figfakelossdistrRandIMGNET

\section{Extended Experimental Results}
\label{appendix:more_results}

In this section, we present additional experimental results as mentioned  in Sec.~\ref{sec:experiments}. 


\subsection{Low-shot Image Classification with Random Initialization}
We present additional results of low-shot image classification for random initialization in Tab.~\ref{tab:real_shots_rand}. From the results, we can draw the same conclusion as in the main paper: applying our proposed Pretrained
Guidance and Real Guidance largely improved the classifier performance in most cases, especially in the low-data
regimes. However, it is interesting to obverse that the dynamic between Pretrained Guidance and Real Guidance are different when starting the networks from random initialization. We believe this can be a direction of future investigations.

\tabrandrealshots

\figfakelsgm
\figfakesgxl
\figreallsgm
\figrealsgxl

\figfakesgIMGNET
\figfakegdIMGNET
\figrealsgIMGNET
\figrealgdIMGNET